\documentclass{article} 
\usepackage{iclr2027_conference,times}

\usepackage{amsmath,amsfonts,bm}

\def\eqref#1{equation~\ref{#1}}

\def\1{\bm{1}}

\DeclareMathAlphabet{\mathsfit}{\encodingdefault}{\sfdefault}{m}{sl}
\SetMathAlphabet{\mathsfit}{bold}{\encodingdefault}{\sfdefault}{bx}{n}

\usepackage{multirow}
\usepackage[hidelinks]{hyperref}
\usepackage{url}
\usepackage{graphicx}
\usepackage{booktabs}
\usepackage{float}
\usepackage[table]{xcolor}
\definecolor{offlinebg}{RGB}{244,248,251}
\definecolor{loopbg}{RGB}{245,250,243}
\usepackage{capt-of}
\iclrfinalcopy

\title{AirPlayground: Benchmarking Aerial Navigation across Diverse Reconstructed Environments}

\author{
{\bfseries
Rangya Zhang$^{1*}$,
Yujing Shang$^{1*}$,
Zhuoyuan Yu$^{1}$,
Mingsheng Li$^{1}$,
Qingsong Yan$^{2}$,
Chao Yan$^{3}$,Yuhang Zhang$^{1\dagger}$,
Mir Feroskhan$^{1\dagger}$
}\\[3pt]
{\normalfont\small
$^{1}$Nanyang Technological University
$^{2}$Wuhan University
$^{3}$Nanjing University of Aeronautics and Astronautics
}
}

\newcommand{\ourname}{AirPlayground}

\begin{document}

\maketitle


\begingroup
\renewcommand{\thefootnote}{\fnsymbol{footnote}}
\footnotetext[1]{Equal contribution.}
\footnotetext[2]{Corresponding authors.}
\endgroup

\begin{abstract}

Large-scale aerial resources provide a foundation for learning navigation across diverse environments. However, existing resources remain limited in providing (1) densely sampled flight demonstrations for learning different policy formulations and (2) broad environment coverage for standardized closed-loop evaluation. To address these limitations, we introduce AirPlayground, a unified benchmark for training and evaluating aerial navigation policies across diverse environments. We first construct 94 photorealistic 3D Gaussian Splatting (3DGS) environments spanning indoor, urban outdoor, and vegetation settings, with aligned collision geometry. Building on these environments, we develop a scene-adaptive pipeline for automated expert flight collection, yielding 94K demonstrations and 46.5M synchronized observation--state--action samples at 50 Hz. We further provide an extensible API that supports training different policy formulations on this shared corpus and evaluating them under unified protocols in the corresponding environments. We benchmark seven baselines spanning four policy paradigms across three navigation settings in seen and unseen environments, and further examine scaling behavior with ACT. Scaling ACT's training corpus from 5 to 70 scenes raises median success rates on seen and unseen environments from $30.6\%$ and $16.8\%$ to $41.2\%$ and $29.4\%$, respectively. Through this shared training and evaluation framework, AirPlayground supports analysis of how offline prediction errors relate to closed-loop outcomes. Our analyses highlight the roles of prediction errors, execution protocols, and scene interactions in closed-loop performance.

\end{abstract}

\vspace{-6pt}
\begin{figure}[H]
\centering
\includegraphics[height=0.25\textheight,keepaspectratio]{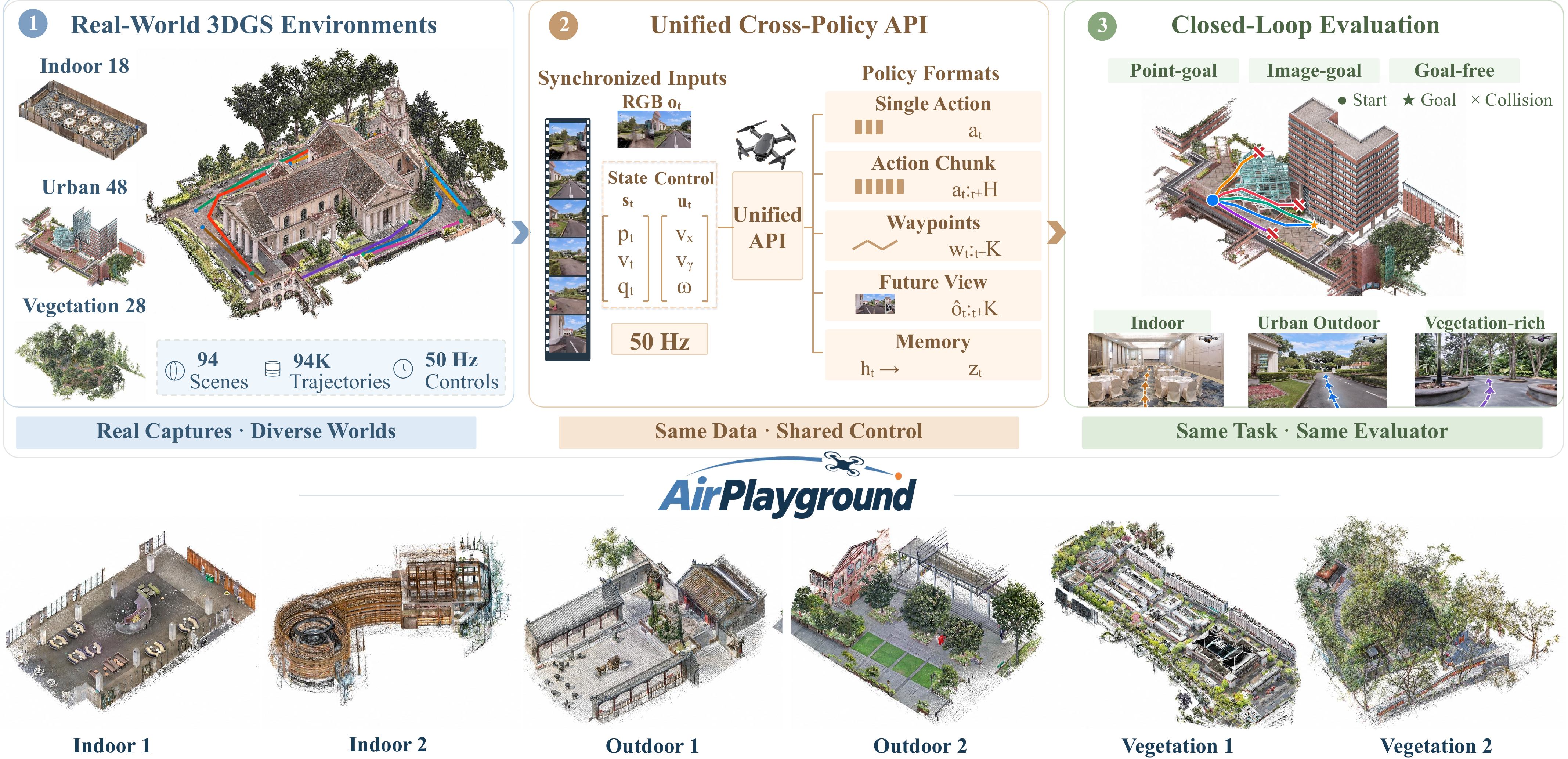}
\caption{\textbf{The AirPlayground benchmark.}
\textit{Top:} (1) 94 real-world scenes are reconstructed as executable 3DGS environments; (2) a unified API adapts the observation--state--action corpus to different policies; and (3) policies are evaluated under closed-loop protocols for point-goal, image-goal, and goal-free navigation.
\textit{Bottom:} representative indoor, urban, and vegetation-rich environments.}

\label{fig:overview}
\end{figure}

\section{Introduction}
Learning-based aerial navigation depends on large-scale, diverse experience to generalize across environments~\citep{zheng2024towards, liu2025citywalker}. Acquiring experience through online interaction can be costly and difficult to scale. Offline demonstrations support scaling policy learning without environment interaction during training~\citep{o2024open}. Yet policies trained on demonstrations learn from expert states, whereas deployment exposes them to states shaped by their actions~\citep{seo2024mitigating}. Small errors can compound through feedback, so accurate imitation does not guarantee reliable navigation. Assessing reliability therefore requires systematic closed-loop evaluation across diverse environments, where policies must respond to the consequences of their actions.

To meet these learning and evaluation needs, aerial navigation research has produced both flight datasets and interactive environments. Large-scale datasets such as CityNav and OpenFly provide aerial trajectories for navigation and vision-language learning~\citep{lee2025citynav,gao2026openfly}, substantially increasing the scale and diversity of reusable flight experience. In parallel, 3DGS-based embodied platforms make reconstructed real-world scenes available as photorealistic, geometrically grounded environments for interactive navigation~\citep{liu2026wanderland,miao2026towards,xia2026habitat}. Recent benchmarks further combine reconstructed environments, collision-aware interaction, and flight trajectories for policy learning and evaluation~\citep{guo2026huge}.

Despite this progress, two limitations remain in the resources for aerial policy learning and evaluation. \textbf{(i) Limited dense flight experience.} Densely sampled observation--action demonstrations remain limited across diverse real-world environments. Dense supervision captures continuous flight control for learning different action representations, while coverage across scenes exposes policies to varied visual conditions and spatial structures. \textbf{(ii) Limited standardized evaluation across diverse environments.} Standardized closed-loop evaluation remains limited in its coverage of environments with diverse appearances and spatial structures. Indoor spaces, urban outdoor areas, and vegetation present different visibility and obstacle configurations, making evaluation across these settings important for assessing how consistently policies perform beyond a narrow set of scenes.

To address these limitations, we introduce AirPlayground, a unified benchmark for aerial navigation that pairs dense flight experience with photorealistic 3DGS reconstructions and executable geometry. AirPlayground comprises \textbf{94 reconstructed scenes} spanning indoor, urban outdoor, and vegetation environments, with \textbf{nearly 100K flight demonstrations} and \textbf{46.5M RGB frames} under 50 Hz observation--action supervision. To scale collection, we develop a scene-adaptive pipeline, reusable across reconstructed scenes, that models navigable space, samples navigation tasks, and generates expert demonstrations. We further introduce an extensible cross-policy API that adapts the shared corpus to each policy's native representation and maps its outputs to a common control interface under unified evaluation protocols. By linking recorded trajectories to executable scene geometry, AirPlayground also enables analysis of how offline prediction errors relate to closed-loop outcomes.

Specifically, AirPlayground makes three main contributions.

\textbf{(i) Dense flight experience paired with executable reconstructions of real-world environments.} We pair dense flight demonstrations with photorealistic reconstructions and aligned navigation and collision geometry for policy learning and matched closed-loop execution.

\textbf{(ii) A unified benchmark across navigation policy paradigms.} Our extensible API supports shared-data policy training and unified closed-loop evaluation across three navigation settings, while preserving each policy’s native representations and inference structure through adapters.

\textbf{(iii) Cross-policy evaluation and deployment diagnosis.} We benchmark seven baselines and analyze offline predictions and closed-loop behavior. These studies show that offline accuracy alone does not fully explain deployment performance and examine how execution horizons and scene geometry affect navigation, demonstrating the benchmark's value beyond performance comparisons.

\section{Related Work}
\paragraph{Aerial Navigation Datasets.}
Prior work has expanded navigation-level supervision from constrained simulated settings toward larger urban environments, richer tasks, and broader scene coverage~\citep{liu2023aerialvln,wang2025towards,lee2025citynav,liu2026indooruav}. More recent efforts scale collection through automated planning and expert trajectory generation~\citep{low2025sous,gao2026openfly,adang2025singer}. Other efforts collect fine-grained flight demonstrations for reactive control and long-horizon navigation~\citep{wang2026uav,zheng2026think}. Despite these advances, temporally dense flight demonstrations remain limited in scale and environmental coverage, constraining the continuous control experience for training diverse navigation policies.

\paragraph{Reconstructed Environments and Navigation Benchmarks.}
Recent embodied-navigation research uses 3D Gaussian Splatting to reconstruct photorealistic real-world environments, providing a real-to-sim bridge for policy learning and evaluation~\citep{chhablani2025embodiedsplat,liu2026navgsim,low2025sous,quach2024gaussian}. To support collision-aware navigation, these representations are augmented with metric geometry, collision models, and navigation structures, enabling interactive navigation across reconstructed environments~\citep{miao2026towards,he2026seeing,liu2026wanderland,xia2026habitat}. Aerial benchmarks also pair 3DGS environments with geometry for collision checking and evaluation: OpenFly spans 18 scenes from multiple rendering sources, while HUGE provides four reconstructed outdoor environments~\citep{gao2026openfly,guo2026huge}. These platforms establish important foundations, but broader coverage of reconstructed indoor, urban outdoor, and vegetation environments is still needed for systematic cross-scene policy evaluation.

\paragraph{Navigation Policy Evaluation.}
Offline evaluation measures prediction errors against expert demonstrations, while interactive aerial benchmarks assess task success and path efficiency through closed-loop rollouts~\citep{xue2026see,lou2026voln}.
Studies in autonomous driving and robot manipulation examine this mismatch, including the predictive value of evaluation metrics and task-critical errors~\citep{wang2026open,huang2026critical}.
This mismatch also depends on how predictions are executed: work on action-chunking policies investigates execution horizons~\citep{zhao2026dynamic}, while studies of aerial agents examine how execution mechanisms affect flight outcomes~\citep{xu2026lightweight}.
However, a systematic understanding of the offline-to-closed-loop gap in aerial navigation remains limited.
By pairing flight demonstrations with executable worlds, AirPlayground supports trajectory replay and policy evaluation in corresponding scenes, providing an experimental basis for tracing offline predictions to their closed-loop consequences.

\section{AirPlayground Benchmark}

\begin{figure*}[t!]
    \centering
    \includegraphics[width=\textwidth]{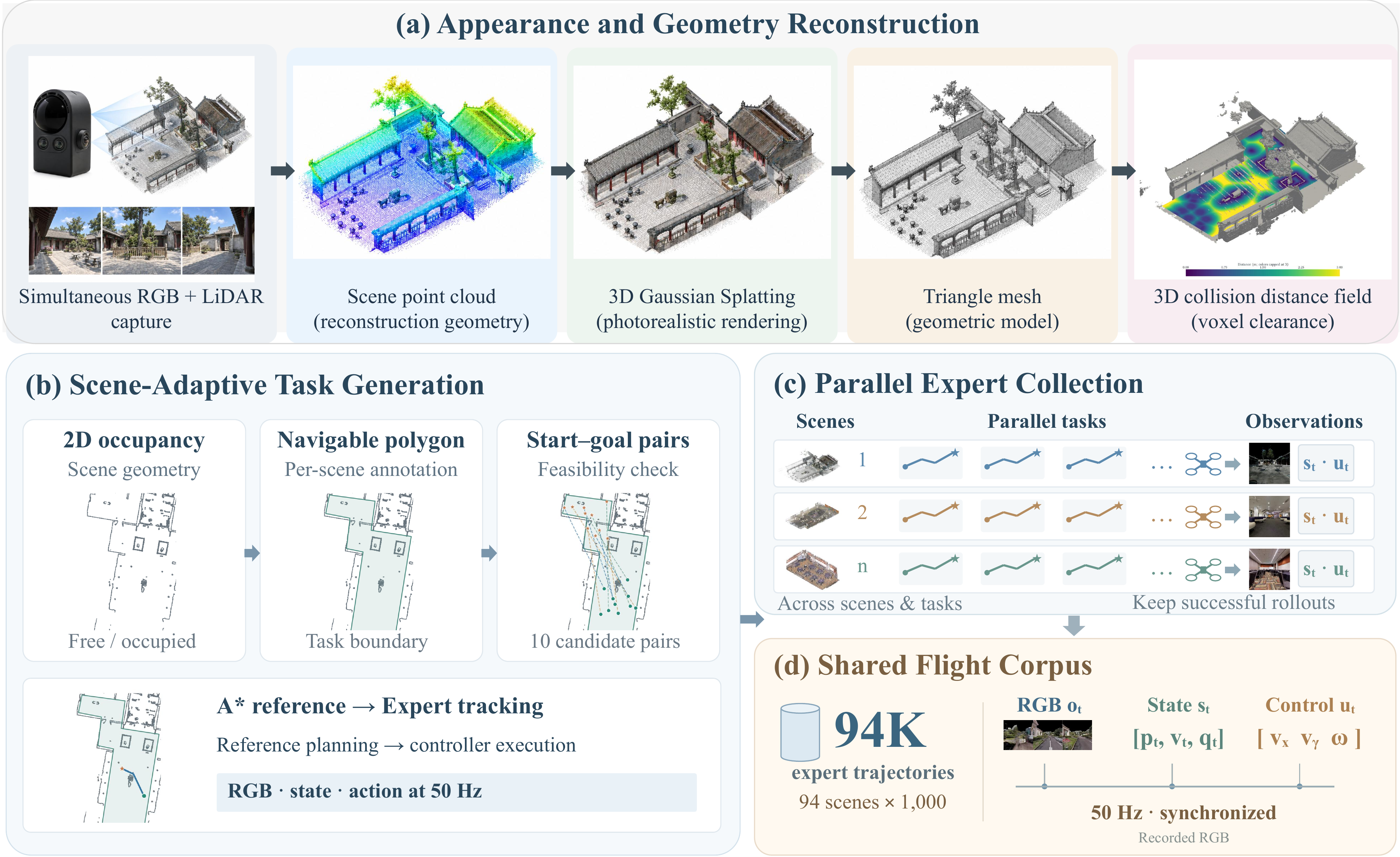}
    \par
    \caption{\textbf{\ourname{} framework.} Real-world captures become executable 3DGS scenes for closed-loop expert collection. A unified cross-policy API supports training on the resulting demonstrations and evaluation in the corresponding scenes.}
    \label{fig:airplayground_framework}
\end{figure*}

This section describes how AirPlayground builds executable environments and dense expert flight experience from real-world captures, and supports policy training and evaluation through a unified interface. Figure~\ref{fig:airplayground_framework} summarizes the construction pipeline, from scene capture and reconstruction to scene-adaptive task generation, expert flight collection, and the resulting shared flight corpus.

\subsection{Scene Collection}
\label{sec:scene_collection}

\paragraph{Capture hardware.}
An operator walks each site at a steady pace carrying an XGRIDS Lixel P1 PortalCam, a handheld scanner that pairs a wide-field LiDAR with four synchronized cameras, two fisheye and two forward-facing. A single walk-through therefore yields both metric geometry and multi-view imagery of the scene (Appendix~\ref{app:collection_space}).

\paragraph{Scene selection and diversity.}
We choose environments spanning three settings with distinct navigation demands: \emph{indoor} spaces such as rooms and corridors, with confined free space and clutter; \emph{urban outdoor} sites such as campuses, streets and landmarks, with regular structures; and \emph{vegetation}-rich environments such as parks, gardens and terrain, with irregular geometry and diffuse obstacle boundaries. Scenes are captured on site or adapted from publicly available assets~\citep{xia2026habitat}, then processed into a common representation (Sec.~\ref{sec:scene_reconstruction}). The corpus contains 94 scenes: 18 indoor, 48 urban outdoor and 28 vegetation. Navigable regions total 75{,}657\,m$^2$, ranging from 14 to 6{,}674\,m$^2$ per scene (median 567\,m$^2$; mean 805\,m$^2$), spanning over two orders of magnitude.

\subsection{Executable Scene Reconstruction}
\label{sec:scene_reconstruction}

\paragraph{Photorealistic 3DGS reconstruction.}
For scenes captured on site, the synchronized LiDAR and camera streams are processed by an end-to-end reconstruction system. It follows the tightly-coupled LiDAR--visual--inertial route to state estimation~\citep{shan2021lvisam,lin2021r2live,zheng2025fastlivo2}, recovering the sensor trajectory and a point map at metric scale. Fisheye images are undistorted, and a 3D Gaussian Splatting model~\citep{kerbl20233dgs} is initialized from the point cloud and optimized under depth supervision on the registered images. The pipeline detects dynamic regions and segments them into masks, and the masked regions are held out of the optimization so that transient content does not enter the reconstruction. Additional assets are brought into the same form, so every scene ends up as a 3DGS asset paired with a metric point cloud in one frame.

\paragraph{Executable geometry.}
A visual asset shows what a scene looks like, but not where a drone may fly or what it would strike. We derive both from the point cloud. A 2D occupancy grid defines the free space for task sampling and path planning. It covers a flight-relevant height band, is restricted to a navigable polygon annotated once per scene, and dilates occupied cells by a per-scene inflation radius so paths keep a margin from obstacles. A 3D clearance field, precomputed once per scene without inflation, answers collision and clearance queries against the robot radius and preserves the vertical structure the 2D grid collapses; GPU queries batch this checking without physics-engine contacts. The 3DGS asset, planning grid, and clearance field share one coordinate frame, so visual observations, task planning, and collision checking refer to the same scene (Appendix~\ref{app:collection_space}).

\subsection{Scene-Adaptive Demonstration Generation}
\label{sec:data_collection}

\paragraph{Scene-adaptive task generation.}
A scene specification records each world's occupancy and inflation settings, navigable polygon and flight altitude. Within the navigable region, the pipeline samples start--goal pairs subject to a minimum separation and a clearance margin from obstacles, keeps a pair only if a planned path connects its endpoints, and resamples otherwise. The initial heading is randomized around the goal direction, so the flown trajectory is not fully determined by the planned path. Where a scene offers little natural obstacle interaction, the pipeline adds box-shaped obstacles in free space to diversify trajectories; the released corpus records their count per episode (Appendix~\ref{app:collection_sampling}).

\paragraph{Closed-loop demonstration collection.}
For each task, A* plans a reference path over the inflated occupancy grid, and greedy line-of-sight shortcutting then shortens it. The expert follows this path in closed loop: at each 50\,Hz control step, a waypoint-tracking expert reads the drone's current simulated state, steers toward the active waypoint, and outputs planar body-frame velocities and a yaw-rate command. Each step synchronously records forward RGB, the UAV state, a local occupancy patch and the executed action. Only successful episodes are retained. Worlds are collected independently and in parallel across GPUs, targeting 1{,}000 successful demonstrations per scene and yielding the corpus reported in Sec.~\ref{sec:dataset_stats}.

\subsection{Unified Cross-Policy API}
\label{sec:cross_policy_api}

Navigation policies differ in observation histories, training targets, and inference and execution mechanisms. AirPlayground handles these differences through training and stateful control adapters, allowing policies to reuse the same flight corpus and evaluation pipeline. The API supports point-goal, image-goal, and goal-free navigation.

Given a trajectory $\tau=\{(o_t,s_t,g_t,a_t)\}_{t=1}^{T}$, a training adapter converts it into the inputs and supervision of policy $m$, $(\mathbf{x}^m,\mathbf{y}^m)=\mathcal{T}_m(\tau)$, constructing observation histories and policy-specific targets, including single-step actions, action chunks, waypoints, or future observations for predictive models. Future observations serve as training supervision, not deployment inputs. For goal-free policies, the adapter withholds the goal input $g_t$ and the relative-goal entries of the state $s_t$.

At deployment, a stateful control adapter maps policy outputs, including actions selected by model-based planning, to body-frame velocity commands, $\mathbf{u}_t=\mathcal{C}_m(\hat{\mathbf{y}}_t^m,s_t;\mathbf{h}_t^m)$ with $\mathbf{u}_t=(v_x,v_y,\dot{\psi})$, where $\mathbf{h}_t^m$ tracks cached predictions and progress through the current plan. It executes each policy's outputs on its own execution schedule, standardizing task inputs and control semantics while preserving method-specific prediction and planning mechanisms. New policies implement these adapters while reusing the task generator, simulator, and evaluator.

\subsection{Resource Statistics and Validation}
\label{sec:dataset_stats}

\paragraph{Corpus scale and schema.}
\ourname{} contains 94 executable worlds and 94{,}000 successful expert trajectories, 1{,}000 per world. Recorded at 50\,Hz, the trajectories comprise 46{,}530{,}557 synchronized RGB--state--action samples, 495 per trajectory on average. They amount to roughly 259 hours and 1{,}800\,km of flight, stored in 5.9\,TB. Each sample pairs a forward $256\times256$ RGB image with a 13-dimensional UAV state that embeds the relative goal, a local occupancy patch, and a continuous body-frame control action $(u_x,u_y,\omega)$ whose translational components are normalized (Appendix~\ref{app:collection_records}). This schema lets every policy formulation in Sec.~\ref{sec:experiments} read from the same trajectories.

\paragraph{Comparison with existing aerial resources.}
Table~\ref{tab:dataset_comparison} compares environment representations, scale, and flight supervision, which differ in granularity and coverage beyond raw trajectory counts. CityNav and OpenFly provide discrete navigation supervision, with OpenFly providing waypoints. UAV-Flow records flight at 5\,Hz in real footage, and FLIGHT at 10\,Hz in synthetic environments, but neither uses 3DGS reconstructions. HUGE-Bench pairs 3DGS with collision geometry across four scenes. \ourname{} combines 94 executable worlds with 94K trajectories and continuous body-frame control supervision at 50\,Hz.

\begin{table*}[t]
\centering
\caption{\textbf{Comparison of aerial navigation resources.} AirPlayground pairs executable 3DGS reconstructions with continuous supervision. ``--'' indicates an absent property or unreported quantity.}
\label{tab:dataset_comparison}
\small

\renewcommand{\arraystretch}{1.10}
\resizebox{\textwidth}{!}{%
\begin{tabular}{llccrrll}
\toprule
& \multicolumn{3}{c}{Environment} & \multicolumn{2}{c}{Scale} & \multicolumn{2}{c}{Flight Experience} \\
\cmidrule(lr){2-4}\cmidrule(lr){5-6}\cmidrule(lr){7-8}
Benchmark & Source & 3DGS & Exec. & Scenes & Traj. & Supervision & Granularity \\
\midrule
AerialVLN~\citep{liu2023aerialvln} & Simulation & -- & \checkmark & 25 & 8.4K & Discrete 4-DoF & $\sim$204 steps/traj. \\
CityNav~\citep{lee2025citynav} & Real-city scans & -- & \checkmark & 2 cities & 32.6K & Discrete actions & -- \\
OpenUAV~\citep{wang2025towards} & Simulation & -- & \checkmark & 22 & 12.1K & 6-DoF trajectory & $\sim$264 steps/traj. \\
OpenFly~\citep{gao2026openfly} & Mixed & \checkmark & Partial & 18 & 100K & Discrete + waypoint & $\sim$35 actions/traj. \\
UAV-Flow~\citep{wang2026uav} & Real + sim. & -- & \checkmark & 3 real & 40.8K & 6-DoF state & 5 Hz \\
FLIGHT~\citep{zheng2026think} & Simulation & -- & \checkmark & -- & 10.8K & Continuous 6-DoF & 10 Hz \\
HUGE-Bench~\citep{guo2026huge} & Digital twins & \checkmark & \checkmark & 4 & 6.2K & RGB + state & $\sim$1\,m spacing \\
\midrule
\rowcolor{violet!8}
\textbf{AirPlayground} & \textbf{Real recon.} & \checkmark & \checkmark & \textbf{94} & \textbf{94K} & \textbf{RGB + state + velocity} & \textbf{50 Hz} \\
\bottomrule
\end{tabular}}
\vspace{1mm}
\end{table*}

\paragraph{Trajectory statistics and validation.}
Planned paths average 19.2\,m (median 13.5\,m, 95th percentile 55.6\,m, maximum 235.1\,m), and episodes last 9.9\,s on average (median 7.1\,s, 95th percentile 27.6\,s). Per-scene frame counts range from 43{,}665 to 2{,}023{,}166. All 94{,}000 released trajectories are successful expert demonstrations, and every one of them is accounted for in storage, with no missing and no orphaned files. All baselines in Sec.~\ref{sec:experiments} use this corpus for training or adaptation and offline evaluation (Appendix~\ref{app:collection_records}).

\begin{figure}[t]
    \centering
    \includegraphics[width=\linewidth]{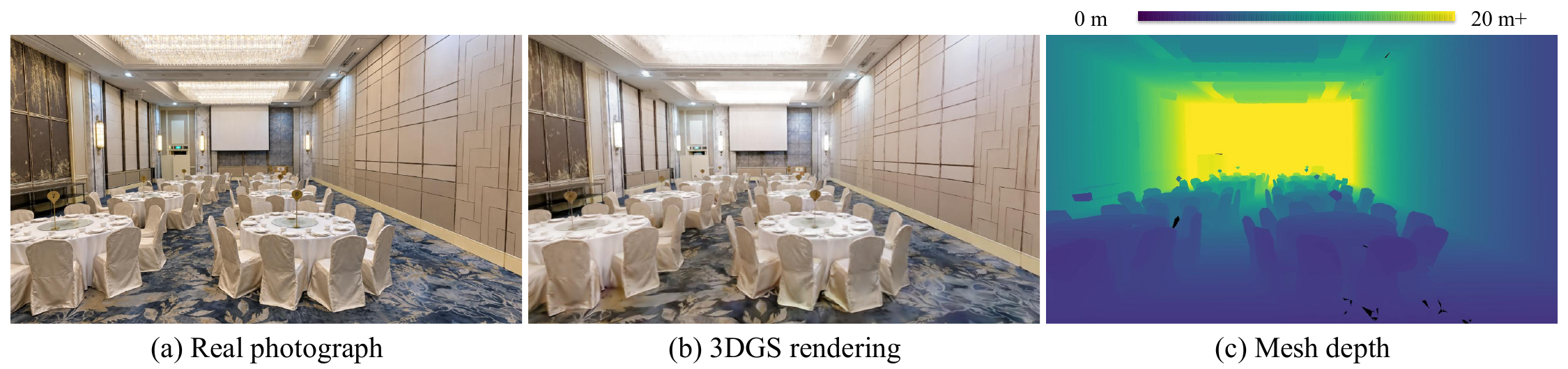}
    \par
    \caption{\textbf{Visual appearance and scene geometry.} (a) Reference photograph, (b) 3DGS rendering, and (c) mesh-derived depth at the matched viewpoint.}
    \label{fig:scene_quality}
\end{figure}

\paragraph{Visual appearance and geometry.}
We evaluate rendering fidelity across 20 scenes with 10 reference images per scene, using camera poses estimated from image correspondences and mesh depth. At $1280\times720$ resolution, the evaluation yields an average PSNR of 19.19\,dB, SSIM of 0.655, and LPIPS (AlexNet) of 0.223. Figure~\ref{fig:scene_quality} illustrates the correspondence between a reference photograph and its 3DGS rendering, preserving the room layout and major furnishings with some differences in fine textures and lighting. Mesh-derived depth depicts the corresponding obstacle surfaces used for collision checking. Together, these results provide evidence of visual fidelity and illustrate how AirPlayground pairs rendered observations with executable scene geometry.

\section{Experiments}
\label{sec:experiments}

We evaluate AirPlayground through three central research questions:
(1) How do aerial navigation policies perform across diverse reconstructed environments under a shared evaluation protocol, and how well do they generalize to unseen scenes?
(2) How does increasing available training experience affect closed-loop performance?
(3) What explains the observed discrepancies between offline prediction accuracy and closed-loop navigation performance?

\subsection{Experimental Setup}
\label{sec:experimental_setup}

\paragraph{Benchmark.}
We split 94 environments into 70 training, 8 validation, and 16 unseen test scenes, with all three categories in each split. Each scene has 1,000 trajectories; training scenes reserve 100 for seen offline evaluation and use 900 for learning. Closed-loop evaluation uses 50 fixed episodes in each of 16 training scenes and all 16 unseen test scenes: 800 per split and 1,600 per policy and setting. Split scores weight scenes equally. Offline-to-closed-loop comparisons use matched scenes.

\paragraph{Task protocol.}
Policies navigate at a fixed flight height using forward RGB, flight state, and task-specific goal inputs, and output planar body-frame velocities and yaw rate. PointGoal supplies a relative target; ImageGoal uses successive views along an A* reference route; Goal-free withholds the target, which is retained only for evaluation. Success requires reaching within 0.45\,m of the goal without a safety violation. Episodes end on success, collision, leaving the navigable region, or a 100\,s timeout; collision checking uses a 0.12\,m body radius.

\paragraph{Baselines.}
We benchmark seven baselines: CNN-BC, ACT~\citep{zhao2023learning}, Diffusion Policy (DP)~\citep{chi2025diffusion}, ViNT~\citep{shah2023vint}, OmniVLA~\citep{hirose2025omnivla}, NWM~\citep{bar2025navigation}, and NavMorph~\citep{yao2025navmorph}. All are trained or adapted on the shared corpus and compared primarily in PointGoal mode under unified protocols while preserving their native prediction and execution mechanisms. Checkpoints and execution intervals are selected on validation scenes and fixed for evaluation (adapter configurations in Appendix~\ref{app:baselines}).

\paragraph{Metrics.}
Offline metrics include average and final displacement errors (ADE/FDE), P95 FDE, and mean absolute endpoint yaw error over a common physical horizon. Closed-loop metrics include success rate (SR), A$^\ast$-SPL, navigation error (NE), collision rate (CR), and out-of-bounds rate (OB). Definitions and additional diagnostics appear in Appendix~\ref{app:metrics}.

\begin{figure*}[t]
    \centering
    \includegraphics[width=\textwidth]{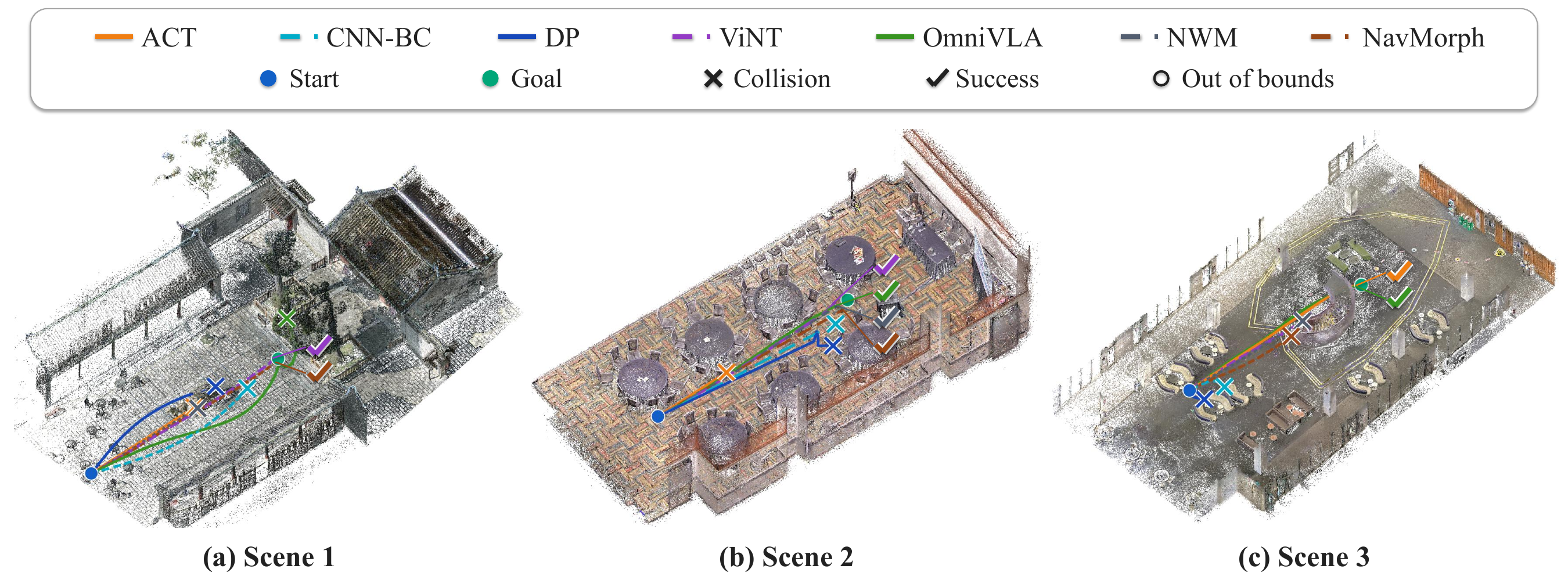}
    \par
    \caption{\textbf{PointGoal rollouts with shared starts and goals in three scenes.}}
    \label{fig:three_scene_navigation}
\end{figure*}

\begin{table*}[t]
\centering
\caption{\textbf{Offline and closed-loop PointGoal performance on seen/unseen scenes.} Offline errors use a 0.2\,s horizon: ADE/FDE/P95 in cm; mean absolute endpoint yaw error in radians. P95 denotes scene-averaged 95th-percentile FDE. Closed-loop protocol: Sec.~\ref{sec:experimental_setup}.}
\label{tab:policy_benchmark}
\scriptsize
\renewcommand{\arraystretch}{1.08}
\resizebox{\textwidth}{!}{%
\begin{tabular}{l *{4}{>{\columncolor{offlinebg}}c} *{4}{>{\columncolor{loopbg}}c} *{4}{>{\columncolor{offlinebg}}c} *{4}{>{\columncolor{loopbg}}c}}
\toprule
& \multicolumn{4}{c}{\cellcolor{offlinebg}Seen Offline}
& \multicolumn{4}{c}{\cellcolor{loopbg}Seen Closed Loop}
& \multicolumn{4}{c}{\cellcolor{offlinebg}Unseen Offline}
& \multicolumn{4}{c}{\cellcolor{loopbg}Unseen Closed Loop} \\
\cmidrule(lr){2-5}\cmidrule(lr){6-9}\cmidrule(lr){10-13}\cmidrule(lr){14-17}
Method & ADE$\downarrow$ & FDE$\downarrow$ & P95$\downarrow$ & Yaw$\downarrow$ & SR$\uparrow$ & A$^\ast$-SPL$\uparrow$ & CR$\downarrow$ & OB$\downarrow$ & ADE$\downarrow$ & FDE$\downarrow$ & P95$\downarrow$ & Yaw$\downarrow$ & SR$\uparrow$ & A$^\ast$-SPL$\uparrow$ & CR$\downarrow$ & OB$\downarrow$ \\
\midrule
CNN-BC & \textbf{0.54} & \textbf{0.81} & \textbf{2.52} & \textbf{0.0022} & 21.0 & 0.21 & 49.0 & 30.0 & \textbf{9.34} & \textbf{16.59} & \textbf{19.75} & \textbf{0.0054} & 15.0 & 0.14 & 47.5 & 37.5 \\
ACT & 1.67 & 3.07 & 16.86 & 0.0096 & 40.8 & 0.41 & 39.2 & 20.0 & 11.84 & 21.55 & 26.68 & 0.0171 & 29.5 & 0.30 & 65.0 & 5.5 \\
DP & 2.88 & 5.38 & 32.98 & 0.0316 & 20.5 & 0.19 & 65.5 & 14.0 & 22.15 & 40.27 & 58.04 & 0.1235 & 11.2 & 0.10 & 48.8 & 40.0 \\
ViNT & 7.11 & 12.81 & 19.85 & 0.0360 & 38.8 & 0.39 & 36.2 & 25.0 & 15.13 & 27.50 & 38.24 & 0.0422 & 33.0 & 0.33 & 64.5 & 2.5 \\
OmniVLA & 4.99 & 8.11 & 15.01 & 0.0454 & 47.5 & 0.47 & 38.8 & 13.7 & 20.64 & 37.95 & 45.58 & 0.0497 & \textbf{40.5} & 0.38 & 59.5 & \textbf{0.0} \\
NWM & 8.70 & 15.81 & 21.60 & 0.0320 & \textbf{50.0} & \textbf{0.50} & \textbf{27.5} & 22.5 & 13.06 & 23.72 & 35.33 & 0.0386 & 38.6 & \textbf{0.39} & 54.3 & 7.1 \\
NavMorph & 2.31 & 3.89 & 13.91 & 0.0347 & 35.5 & 0.35 & 59.5 & \textbf{5.0} & 11.98 & 21.64 & 28.38 & 0.0180 & 22.5 & 0.22 & \textbf{46.3} & 31.2 \\
\bottomrule
\end{tabular}}
\end{table*}

\subsection{Unified Policy Benchmark}
\label{sec:benchmark}

\paragraph{Unified evaluation reveals complementary policy strengths.}
Evaluating different policies on the same flight corpus and task reveals complementary strengths across metrics and scene splits (Table~\ref{tab:policy_benchmark}). NWM achieves the highest seen-scene SR and A$^\ast$-SPL with the lowest CR, while NavMorph attains the lowest OB. On unseen scenes, OmniVLA leads in SR and OB, NWM in A$^\ast$-SPL, and NavMorph in CR. Comparable success rates can mask execution behaviors: NWM and OmniVLA achieve seen-scene SRs of 50.0\% and 47.5\%, yet NWM exhibits lower control variation and path turning (ControlVar 0.003; Turn/m 0.120) than OmniVLA (0.540; 3.919). Across navigation settings, ImageGoal with successive image subgoals yields higher SR than PointGoal for three of seven baselines, while withholding goal input lowers SR for all seven (Table~\ref{tab:app_goal_modes}), revealing policy-dependent responses to goal guidance. Figure~\ref{fig:three_scene_navigation} illustrates paths and outcomes in three scenes.

\paragraph{Better offline prediction does not always yield better navigation.}
CNN-BC achieves the lowest seen-scene ADE and FDE but only 21.0\% SR.
Averaging seen and unseen scores for each policy, ADE and FDE have Spearman correlations of $\rho=0.29$ with SR, opposite to the expected negative relationship.
Their correlations with CR are weak ($\rho=0.18$), while FDE correlates negatively with OB ($\rho=-0.43$), contrary to expectation.
Thus, lower displacement error does not consistently indicate better closed-loop performance, motivating examination of deployment-relevant error structure.

\begin{figure*}[t!]
\centering

\begin{minipage}[t]{0.48\textwidth}
    \centering
    \includegraphics[width=\linewidth]{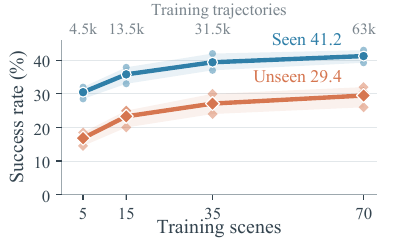}
    \par
    \captionof{figure}{\textbf{ACT data scaling.} Three training seeds evaluated on five seen and five unseen scenes, with 100 fixed episodes per scene.}
    \label{fig:act_scaling}
\end{minipage}
\hfill
\begin{minipage}[t]{0.48\textwidth}
    \centering
    \includegraphics[width=\linewidth]{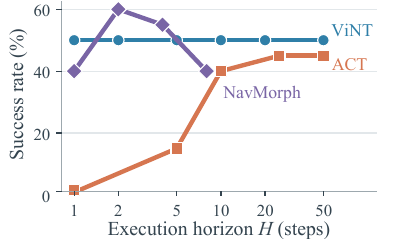}
    \par
    \captionof{figure}{\textbf{Execution-horizon sensitivity.} Averaged SR over 5 seen and 5 unseen scenes; 50 episodes per scene, unchanged checkpoints.}
    \label{fig:execution_horizon}
\end{minipage}

\end{figure*}

\paragraph{Scaling flight experience improves closed-loop generalization.}
We study scaling using ACT trained on nested subsets of 5, 15, 35, and 70 scenes, with 900 trajectories per scene.
We keep the optimization settings and epochs fixed and train each configuration with three seeds.
The scaling protocol uses 100 fixed episodes per scene in the five seen scenes shared by all training subsets and five fixed unseen scenes (500 episodes per split and seed), with identical evaluation tasks across training scales and seeds.
In Figure~\ref{fig:act_scaling}, median success rates across the three seeds rise over the four scales on both splits: unseen-scene SR from 16.8\% to 29.4\%, seen-scene SR from 30.6\% to 41.2\%.

\begin{figure}[t]
\centering
\includegraphics[width=0.98\textwidth]{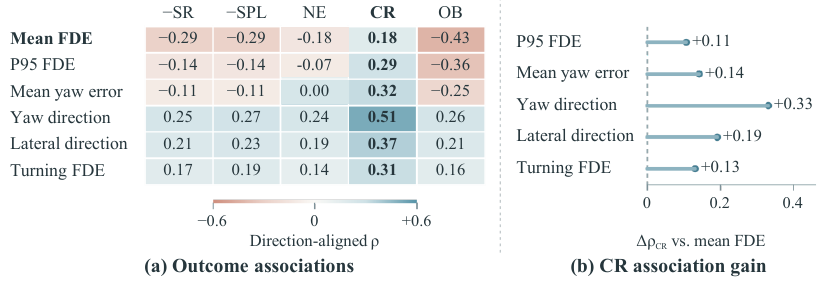}
\caption{\textbf{Offline diagnostics and closed-loop outcomes.} (a) Direction-aligned Spearman correlations: positive correlations link errors to worse outcomes. (b) Signed correlation differences for CR relative to mean FDE. Correlations use seven scores, averaging seen/unseen splits (Appendix~\ref{app:offline_diagnostics}).}
\label{fig:offline_deployment}
\end{figure}

\subsection{From Prediction to Consequence}
\label{sec:prediction_consequence}

Sec.~\ref{sec:benchmark} shows that aggregate offline errors do not consistently track closed-loop outcomes. Using paired demonstrations and executable scenes, we examine what error structure, execution timing, and local geometry reveal about this mismatch.

\paragraph{Error structure matters for deployment diagnosis.}
\label{sec:offline_signals}
We relate fine-grained offline errors to aggregated closed-loop outcomes across policies (Fig.~\ref{fig:offline_deployment}(a)). Mean FDE shows a weak positive association with CR ($\rho=0.18$), whereas yaw direction error, lateral direction error, and turning-subset FDE show stronger associations ($\rho=0.51$, $0.37$, and $0.31$, respectively). Figure~\ref{fig:offline_deployment}(b) shows signed correlation differences relative to mean FDE, reaching $+0.33$ for CR. These results suggest that directional and maneuver-conditioned errors retain collision-related information beyond mean FDE, highlighting their potential for offline safety diagnosis.

\paragraph{Execution horizon changes more than query frequency.}
\label{sec:execution}
Offline prediction scores do not capture how predictions are consumed during interaction. We vary the steps $H$ between model queries across five seen and five unseen scenes, using 50 episodes per scene and the same checkpoints and tasks across horizons. For ACT, each query replaces the previous action chunk, and only its first $H$ controls are executed, without temporal ensembling. Figure~\ref{fig:execution_horizon} shows policy-dependent responses: ACT's SR rises from 0\% at $H=1$ to 45\% at longer horizons, NavMorph performs best at an intermediate horizon, and ViNT's SR remains unchanged across tested settings. More frequent policy queries therefore do not necessarily improve navigation.

\begin{figure}[t]
    \centering
    \includegraphics[width=\linewidth]{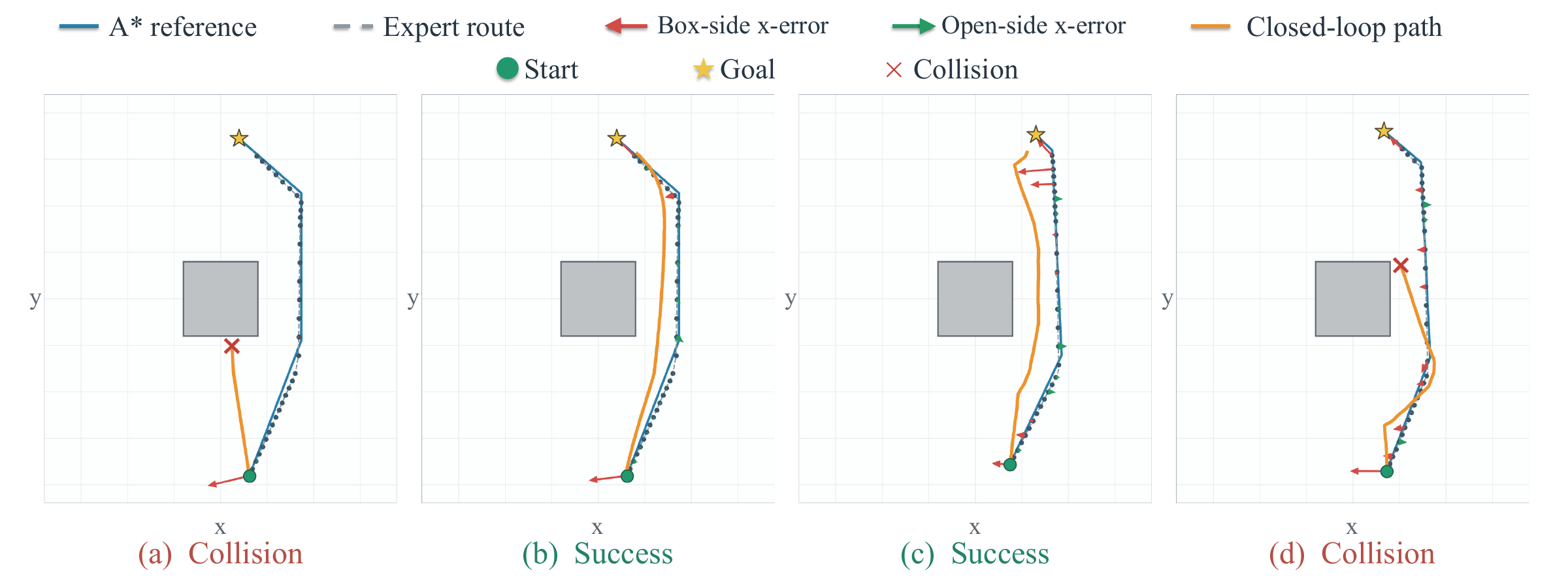}
    \par
    \caption{\textbf{Offline prediction errors and closed-loop trajectories around an obstacle.}}
    \label{fig:geometry_clearance_cases}
\end{figure}

\begin{table}[t]
\centering
\caption{\textbf{Outcomes by predicted clearance change.} Group-mean distances; counts out of $N$.}
\label{tab:geometry_clearance}
\small
\resizebox{\linewidth}{!}{%
\begin{tabular}{lcccccc}
\toprule
Change & $N$ & FDE (cm) & Expert $\rightarrow$ Pred. (cm) & Closed-loop min. (cm) & Collision & Local pass \\
\midrule
Decrease & 43 & 8.4 & 25.6 $\rightarrow$ 18.9 & 5.2 & 16/43 (37.2\%) & 25/43 (58.1\%) \\
Stable & 76 & 7.9 & 26.2 $\rightarrow$ 26.5 & 10.8 & 16/76 (21.1\%) & 56/76 (73.7\%) \\
Increase & 61 & 8.8 & 25.9 $\rightarrow$ 32.1 & 15.7 & 8/61 (13.1\%) & 49/61 (80.3\%) \\
\bottomrule
\end{tabular}}
\end{table}

\paragraph{Local geometry modulates the safety cost of action deviations.}
\label{sec:geometry}
Prediction error measures deviation from an expert trajectory but not its effect on obstacle clearance. Figure~\ref{fig:geometry_clearance_cases}(a--b) shows opposite outcomes under similar layouts and visually similar offline error patterns; in (d), the rollout initially follows the reference but later collides. These cases highlight the importance of where deviations occur relative to obstacles. We group 180 local obstacle-interaction samples by the difference between predicted and expert minimum clearance over the same prediction window, using $\pm2$\,cm as the stable range. Local passage requires reaching the predefined exit within
5\,s without collision or leaving the navigable region. Despite similar group-average prediction errors and expert clearances, the clearance-decrease group has lower closed-loop minimum clearance (5.2 versus 15.7\,cm), a higher reference-obstacle collision rate (37.2\% versus 13.1\%), and a lower local passage rate (58.1\% versus 80.3\%) than the increase group (Table~\ref{tab:geometry_clearance}). These associations highlight safety consequences not directly described by displacement error, motivating geometry- and collision-aware policy learning.

\section{Conclusion}
\label{sec:conclusion}

We introduce AirPlayground, a unified benchmark for aerial navigation that pairs dense flight experience and scene geometry with executable real-world reconstructions. By adapting shared trajectories and unifying deployment controls, our cross-policy API enables different policy formulations to learn from the same flight corpus and undergo standardized evaluation across diverse environments. We benchmark seven baselines spanning four policy paradigms across three navigation settings in seen and unseen scenes, while providing a unified interface for additional policies. Beyond measuring task performance, our analyses reveal an offline-to-closed-loop mismatch involving error structure, execution protocols, and interactions with scene geometry. Together, these analyses demonstrate how AirPlayground combines standardized performance comparison with deployment diagnosis in the corresponding environments. We will publicly release the flight dataset, cross-policy API, and benchmark tools. We hope AirPlayground will serve as a foundation for aerial policy learning and evaluation, helping researchers train policies on dense flight experience, diagnose behavior in executable worlds, and explore geometry- and collision-aware learning.

\section*{AI Use Statement}
We used generative AI tools solely to improve the grammar and clarity of text originally drafted by the authors. These tools were not used for research ideation, method design, code implementation, experiment execution, or analysis and interpretation of results. The authors reviewed all AI-assisted text to ensure that it accurately reflects their own work and findings. We take full responsibility for the final content of this paper, including all text, claims, and artifacts.

\bibliography{iclr2027_conference}
\bibliographystyle{iclr2027_conference}

\appendix
\section{Demonstration Collection Protocol}
\label{app:collection}

\ourname{} generates navigation demonstrations by sampling tasks in reconstructed scenes, planning reference paths, and executing a feedback controller in the simulator. The collection unit is a scene-specific PointGoal episode. Each episode contains observations and control commands from the executed flight, together with the task configuration and reference path. The following description details the base collection pipeline and separately documents its optional diversity mechanisms. Table~\ref{tab:collection_defaults} lists the reference implementation settings; scene configuration files and launch arguments determine the settings of an individual collection run.

\subsection{Scene Configuration and Navigable Space}
\label{app:collection_space}

Each scene is associated with a rendered 3DGS asset, geometric point clouds, and an annotated navigable polygon. A scene configuration specifies the geometric transformation, the height interval used for obstacle extraction, the obstacle inflation radius, and the initial flight height or height range. The collection code resolves these scene-specific settings before constructing the planning grid, so a scene's planning grid is built the same way in every run. Settings that vary between collection runs, such as the number of virtual obstacles, are supplied on the command line.

\begin{figure}[t]
\centering
\includegraphics[width=\linewidth]{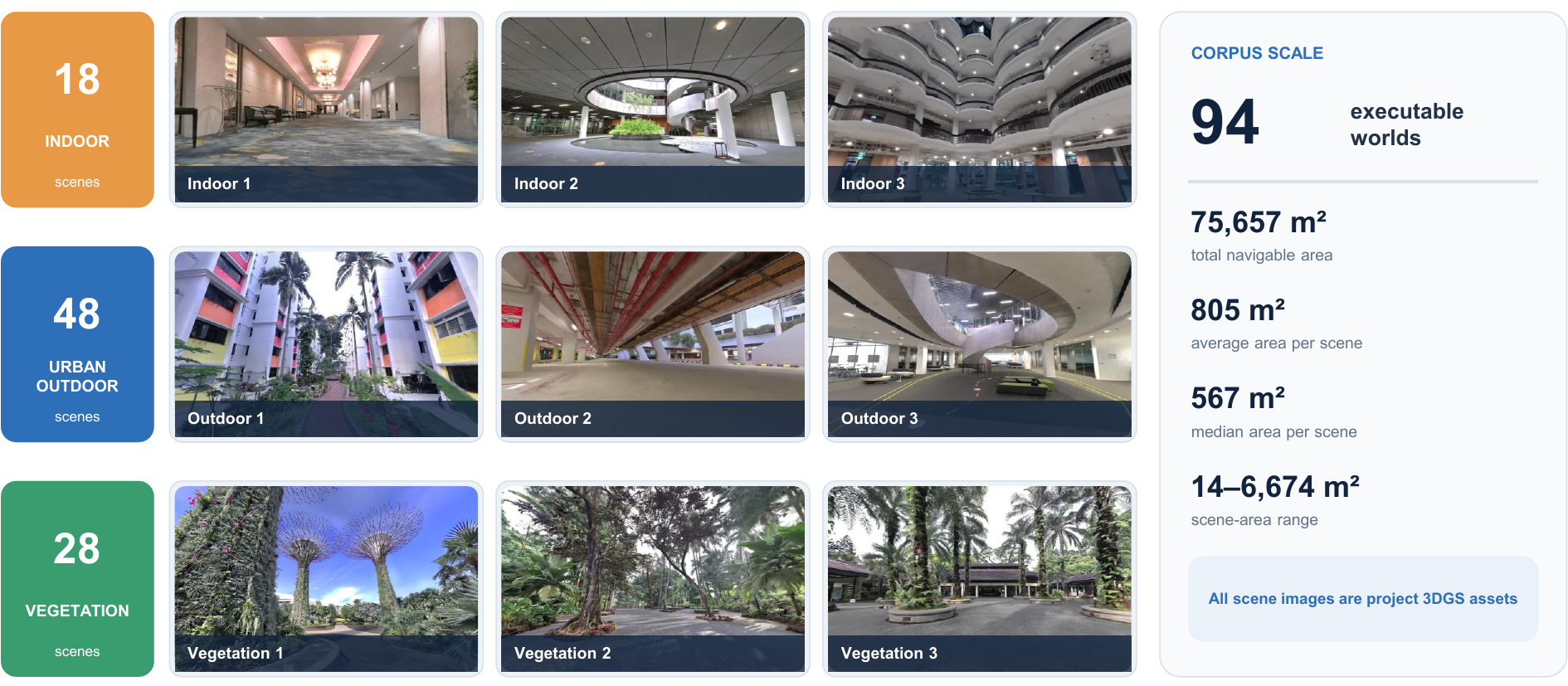}
\caption{\textbf{Scene diversity and corpus scale.}
Representative views across indoor, urban outdoor,
and vegetation environments.}
\label{fig:app_scene_diversity}
\end{figure}

\paragraph{Capture device.}
Scenes captured on site are recorded with an XGRIDS Lixel P1 PortalCam. Its LiDAR is a Class~1 940\,nm unit with a $180^\circ\!\times\!180^\circ$ field of view and a rate of 856{,}000 points per second, ranging to 30\,m against 10\% reflectivity and to 60\,m against 90\%. Its camera array comprises two fisheye cameras ($200^\circ\!\times\!200^\circ$) and two forward-facing cameras ($100^\circ\!\times\!85^\circ$), each at $4000\times3000$ resolution. The unit carries no RTK receiver.

\paragraph{Scene assets and frames.}
A scene ships as a USD package holding the renderable asset together with a triangle mesh of the same reconstruction, alongside a separate point-cloud file. Because all of these come from one reconstruction, they already share a frame; the collection code applies a single fixed map transform to bring them into the simulation frame, and no pairwise registration between appearance and geometry is performed. The packaged mesh is exposed to the environment as a ray-cast target for depth queries. The clearance field described below is what supplies collision geometry, and it is derived from the point cloud rather than from that mesh.

\paragraph{Occupancy construction.}
Point-cloud coordinates are scaled, rotated, and translated into the simulation frame. Points within the configured height interval are projected onto a horizontal grid with cell width $\delta$. A cell is initially occupied when its point count reaches the configured threshold; an optional vertical-extent threshold further filters the projected points. Small connected components are removed before obstacle inflation. Occupied cells are then dilated by a discrete circular neighborhood with radius $\lceil r_{\mathrm{inf}}/\delta\rceil$, where $r_{\mathrm{inf}}$ is the configured inflation distance. Finally, cells whose centers lie outside the navigable polygon are marked occupied. The resulting grid defines the feasible domain for sampling and path planning.

The planning grid describes horizontal traversability. A feasible grid path is followed by an actual closed-loop rollout before the episode is accepted, so grid feasibility and successful execution are separate checks in the collection procedure.

\paragraph{Collision distance field.}
Because the planning grid is horizontal, it cannot distinguish an obstacle that is dangerous only at ceiling height from one that is dangerous at flight height. A separate 3D clearance field supplies that distinction. We first rebuild the footprint above without inflation, then voxelize the height-filtered, polygon-cropped point cloud on the same horizontal grid extended by a vertical axis at the same resolution, and keep only voxels whose $(x,y)$ column survives the cleaned footprint. This retains the two-dimensional denoising while recovering height structure, which a 3D connected-component filter applied directly would fragment. A Euclidean distance transform over the remaining free voxels, scaled by the voxel size, yields the field, stored with its origin and resolution.

The bake applies no inflation. At run time a query point is mapped to its voxel by flooring, the stored clearance is read in a vectorized gather on the simulation device, and the robot radius is subtracted; a non-positive result signals collision. Points outside the baked volume are treated as free, and departures from the scene are handled separately by out-of-bounds and height checks. The bake reuses the simulator's map transform so that the clearance field, the planning grid and the rendered asset share one frame; this correspondence has to be maintained whenever a scene overrides that transform.

\subsection{Task Sampling and Initialization}
\label{app:collection_sampling}

\paragraph{Candidate generation.}
The sampler draws horizontal positions within the polygon's bounding box and rejects positions outside the polygon. For endpoint sampling, it further dilates the planning occupancy by an endpoint-clearance margin. Candidate positions are mapped to nearby free cells in this sampling mask and checked again for polygon membership. Duplicate cells are removed, yielding a set of up to $N$ candidate locations.

\paragraph{Start--goal selection.}
For candidate locations $\{\mathbf{c}_i\}_{i=1}^{N}$, the sampler constructs eligible unordered pairs
\begin{equation}
\mathcal{E}=\left\{(i,j):i<j,\;
\|\mathbf{c}_i-\mathbf{c}_j\|_2\geq\max(d_{\min},2\delta)\right\},
\end{equation}
where $d_{\min}$ is the minimum start--goal separation. A pair is selected uniformly from this set, and its start and goal roles are randomly exchanged. If sampling or subsequent planning fails, the collector resamples the task. The path-length distribution this produces is measured on the collected episodes (Sec.~\ref{sec:dataset_stats}).

\paragraph{Initial pose.}
The robot and goal share an episode-specific initial height drawn from the scene configuration. Planning operates in the horizontal plane, and the action interface supplies horizontal body-frame velocity and yaw rate, with zero commanded body-frame vertical velocity. The robot is initialized at rest. In the default goal-facing initialization, its yaw points toward the goal with a uniformly sampled offset in $[-45^{\circ},45^{\circ}]$. This produces variation in the initial heading while keeping the task endpoints fixed.

\paragraph{Virtual obstacle placement.}
Beyond the geometry recovered from the scene itself, episodes may include virtual box obstacles placed in the free space of the planning grid. The number of virtual obstacles is fixed within a collection run. Given the sampled start and goal, their centers are drawn by rejection sampling over free grid cells: a candidate is accepted when its box footprint, expanded by a planning clearance, contains no occupied cells, and when it lies at least a configured distance from the start, from the goal, and from every previously accepted virtual obstacle. Sampling stops once the requested number of centers is accepted; an episode whose virtual obstacles cannot be placed within the attempt budget is rejected and resampled. Each accepted obstacle is assigned a random yaw, and its footprint, expanded by a planning clearance, is written into the planning occupancy grid before path search. Episode metadata records the number of placed obstacles and their positions; in the released corpus, 65{,}000 of the 94{,}000 episodes (69.1\%) contain virtual obstacles.

\subsection{Reference Path Planning}
\label{app:collection_planning}

A* searches an eight-connected occupancy grid. Horizontal and vertical moves have unit cost, diagonal moves have cost $\sqrt{2}$, and the heuristic is Euclidean distance in grid coordinates. A diagonal transition is allowed only when both adjacent axial cells are free, preventing corner cutting during grid search. Endpoints are associated with nearby free cells within a bounded snapping radius, and the search runs under a bounded expansion budget; a failed endpoint association, an exhausted budget, or an unsuccessful search causes the task to be rejected.

After search, a greedy shortcutting pass reduces the number of intermediate waypoints. Starting from the current anchor, it selects the farthest subsequent waypoint whose connecting segment passes the grid's discrete free-space check, and repeats this operation until the goal is reached. Grid points are converted back to scene coordinates, and the path endpoints are set to the sampled start and goal. For the resulting polyline $P=(\mathbf{p}_0,\ldots,\mathbf{p}_K)$, the reference length is
\begin{equation}
L(P)=\sum_{k=0}^{K-1}\|\mathbf{p}_{k+1}-\mathbf{p}_k\|_2.
\end{equation}
The collector stores both the waypoint sequence and its length. Across the released corpus, the mean ratio of executed path length to straight-line start--goal distance is 1.06.

\subsection{Closed-Loop Expert Execution}
\label{app:collection_control}

\paragraph{Waypoint tracking.}
The expert maintains an index into the reference polyline. At each control step, it advances past intermediate waypoints within a capture radius $r_{\mathrm{wp}}$ of the current horizontal position. The active waypoint is lifted to the robot's current height and transformed into the body frame using the current pose. Let $\mathbf{e}_t=(e_{x,t},e_{y,t})$ denote its horizontal body-frame displacement and $d_{g,t}$ the horizontal distance to the final goal. The base controller uses the goal-dependent speed fraction
\begin{equation}
b_t=\min\left(1,\max\left(0.25,\frac{d_{g,t}}{r_{\mathrm{slow}}}\right)\right)
\end{equation}
and produces normalized translational commands
\begin{equation}
\mathbf{u}_t=b_t\frac{\mathbf{e}_t}{\|\mathbf{e}_t\|_2},\qquad
\omega_t=\operatorname{clip}\left(k_\psi\operatorname{wrap}\left(\operatorname{atan2}(e_{y,t},e_{x,t})\right),-\omega_{\max},\omega_{\max}\right).
\end{equation}
Here, $\operatorname{wrap}$ maps angles to $[-\pi,\pi)$, and a near-zero waypoint displacement produces a zero action. Translational components are clipped to the action bounds. The controller recomputes these commands from the current simulated pose at every step; the reference path remains fixed during a base rollout.

\paragraph{Action semantics.}
The stored action is $\mathbf{a}_t=(u_{x,t},u_{y,t},\omega_t)$. The wrapper inserts a zero vertical component and passes the command to the simulator. With configured translational scale $v_{\mathrm{scale}}$, the simulator applies
\begin{equation}
\mathbf{v}^{b}_t=v_{\mathrm{scale}}
\frac{(u_{x,t},u_{y,t},0)}{\max(1,\|(u_{x,t},u_{y,t},0)\|_2)}.
\end{equation}
Yaw rate is passed directly as the body-frame angular velocity about the vertical axis. The simulator rotates the linear and angular velocities into the world frame and writes them to the robot state. Thus, the first two stored action components require the configured velocity scale to recover physical units. This is a velocity-controlled simulation interface; the demonstrations describe navigation under that interface.

\paragraph{Episode acceptance.}
An episode is accepted when the environment reports success or the robot enters the collector's horizontal goal-radius threshold. Episodes that terminate without success are excluded from the standard demonstration output, and the task is resampled. Collection continues until the requested number of successful demonstrations has been saved.

Discarded rollouts are recorded in the collection logs. The most recent collection run and its resumption together saved 65{,}000 successful episodes, of which 64{,}000 enter the released corpus; 1{,}332 rollouts terminated without reaching the goal and two aborted on a simulator error, so 98.0\% of the rollouts flown in that run were kept. These counts cover rollouts only; tasks rejected at sampling or planning time are resampled before any rollout and are not logged as failures.

\begin{table}[t]
\centering
\small
\caption{Reference settings of the collection implementation. Scene files and launch arguments must be consulted when reproducing a particular collection run.}
\label{tab:collection_defaults}
\begin{tabular}{ll}
\toprule
Parameter & Reference value \\
\midrule
\multicolumn{2}{l}{\emph{Navigable space and task sampling}} \\
Planning-grid cell width $\delta$ & $0.10$\,m \\
Endpoint candidates $N$ & $128$ \\
Minimum endpoint separation $d_{\min}$ & $2.0$\,m \\
Additional endpoint-clearance margin & $0.40$\,m \\
Endpoint distance-quantile filter & $0.0$ \\
Task resampling attempts & $20$ \\
\midrule
\multicolumn{2}{l}{\emph{Virtual obstacles}} \\
Box footprint & $0.8\times0.8$\,m \\
Planning clearance & $0.20$\,m \\
Clearance from start & $0.80$\,m \\
Clearance from goal & $1.00$\,m \\
Minimum spacing & $1.20$\,m \\
Placement attempts & $2048$ \\
\midrule
\multicolumn{2}{l}{\emph{Planning and control}} \\
Planner snapping radius & $0.80$\,m \\
Waypoint capture radius $r_{\mathrm{wp}}$ & $0.18$\,m \\
Goal slowdown radius $r_{\mathrm{slow}}$ & $0.80$\,m \\
Collector goal radius & $0.45$\,m \\
Yaw proportional gain $k_\psi$ & $2.5$\,s$^{-1}$ \\
Yaw-rate bound $\omega_{\max}$ & $1.0$\,rad/s \\
Velocity scale $v_{\mathrm{scale}}$ & $2.0$\,m/s \\
Simulation step / control interval & $0.01$\,s / $0.02$\,s \\
\midrule
\multicolumn{2}{l}{\emph{Recording}} \\
RGB resolution & $256\times256$ \\
Local occupancy shape & $64\times64$ \\
Local occupancy resolution & $0.10$\,m \\
\bottomrule
\end{tabular}
\end{table}

\subsection{Recorded Modalities and Episode Storage}
\label{app:collection_records}

The collector records the current observation and selected action before advancing the simulator. A record contains forward RGB, a 13-dimensional state vector, a local occupancy patch, and a three-dimensional action. RGB is stored as an unsigned 8-bit array in height--width--channel order. State and action arrays use 32-bit floating-point values.

The state vector concatenates body-frame linear velocity (three entries), body-frame angular velocity (three), the goal position relative to the robot in the body frame (three), and the world-frame orientation quaternion in $(w,x,y,z)$ order (four). The relative goal is therefore embedded in the state array. The local occupancy patch is centered at the robot and aligned with its yaw, with forward along decreasing row indices and body-left along increasing column indices. Occupied cells have value one, free cells have value zero, and samples outside the grid are marked occupied. Its cell size is recorded per episode; at the reference resolution of $0.10$\,m the $64\times64$ patch covers $6.4\times6.4$\,m. It is computed from scene geometry and constitutes geometric supervision; its presence in the corpus does not prescribe which inputs a benchmark policy receives.

Variable-length episodes are stored individually or concatenated along the time dimension in compressed NPZ shards. Shards include episode identifiers, start offsets, lengths, and success flags, so episode boundaries can be recovered exactly. Companion JSON metadata records task endpoints, initial pose information, planning configuration, reference waypoints, path length, episode length, and available camera and collection settings. Optional diagnostic videos are stored separately from the training arrays.

\paragraph{Integrity checks.}
The released corpus is reconciled against its own index in both directions. Every one of the 94{,}000 episodes recorded in metadata resolves to a shard file that is present, non-empty and opens as a valid archive; every one of the 940 shard files is referenced by that metadata; and no missing or orphaned files remain on either side. Shard metadata reports the expected count of successful episodes throughout, and the frame totals reported above are summed from the per-episode lengths it records. Beyond this reconciliation, the corpus has been consumed end to end: the seven baselines of Sec.~\ref{sec:experiments} read these shards directly for training and offline evaluation, so the recorded arrays load, stay aligned across modalities, and support policy learning.

\subsection{Behavior Diversity Mechanisms}
\label{app:collection_diversity}

Beyond the base episodes, the collector provides mechanisms that extend task and behavior coverage. All of them record into the same observation--action schema as base episodes.

\paragraph{Obstacle- and corner-aware speed.}
An optional multiplier reduces speed according to nearby occupied cells and upcoming path bends. Clearance is measured from the center of the local occupancy patch to the closest occupied cell in a forward-facing region that includes a small rear margin. A clipped clearance ratio determines one speed factor. A second factor decreases as the robot approaches a waypoint whose turning angle reaches a specified threshold. The controller uses the smaller factor, bounded below by the configured minimum speed fraction, together with the base goal slowdown.

\paragraph{Near-obstacle starts.}
The hard-start sampler selects free cells from a configurable clearance band around obstacles. It samples a separate feasible goal and varies the initial heading using uniform, obstacle-facing, or goal-facing modes. The sampled start must remain inside the scene polygon, and the resulting task must pass the same planning and execution checks as a base episode.

\paragraph{Perturbation and recovery.}
During selected episodes, the collector temporarily holds an earlier action or adds a bounded horizontal command bias. After the disturbance ends, it replans from the current robot position to the original goal and resumes waypoint tracking. The action array stores the command actually selected during the disturbance, while event metadata additionally records the contemporaneous expert commands, event timing, and replanning outcome. These fields distinguish disturbance segments from subsequent expert recovery when constructing training samples.

\paragraph{Planning and camera variation.}
An episode-level inflation offset changes the planning margin, while the local occupancy observation continues to use the base geometry representation. This keeps observation geometry consistent across planning-margin draws. Camera variation changes horizontal field of view at episode boundaries while retaining the square RGB resolution. The collector records the drawn settings and available runtime camera metadata for later analysis.

\subsection{Collection across Scenes}
\label{app:collection_parallel}

Scene collection processes operate independently, each loading its own scene assets, sampling tasks, executing the expert, and writing episode shards. An orchestration layer runs one process per scene and packs several processes onto each GPU, so that a single launch fills every available GPU up to a configured per-GPU process count and queues the remaining scenes. Collection writes to fast local storage and promotes a scene to durable network storage only once its shard metadata reports the requested episode count and every archive opens as a valid file, which keeps the step loop off the network-storage path. A reference count derived from already-collected scenes sets the per-scene target, a resumable staging layout supports continuation after interruptions, and a watchdog frees the GPU slot of a collector that reports a large, configurable number of attempts without a success. Failed task attempts are resampled within a scene, and process-level failures do not halt the remaining queue. The common recording schema permits outputs from independently collected scenes to be aggregated for policy training.

\section{Experimental Details}
\label{app:experiments}

\subsection{Evaluation Protocol and Metrics}
\label{app:metrics}
\label{app:eval_protocol}
\suppressfloats[t]
\providecommand{\AEheadingfont}{\ifdefined\XeTeXversion\fontfamily{lmr}\selectfont\fi}

\paragraph{{\AEheadingfont\textbf{Data split.}}}
All policy training follows the dataset partitions in Table~\ref{tab:app_eval_scenes}, which reports scene counts, stored trajectories, and scene polygon areas. Each partition covers indoor, urban outdoor, and vegetation environments. Each training scene provides 900 trajectories for training and 100 held-out trajectories for seen offline evaluation. Validation scenes are used for checkpoint selection; test scenes are excluded from training and model selection.

\begin{figure}[t]
\centering
\includegraphics[width=\linewidth]{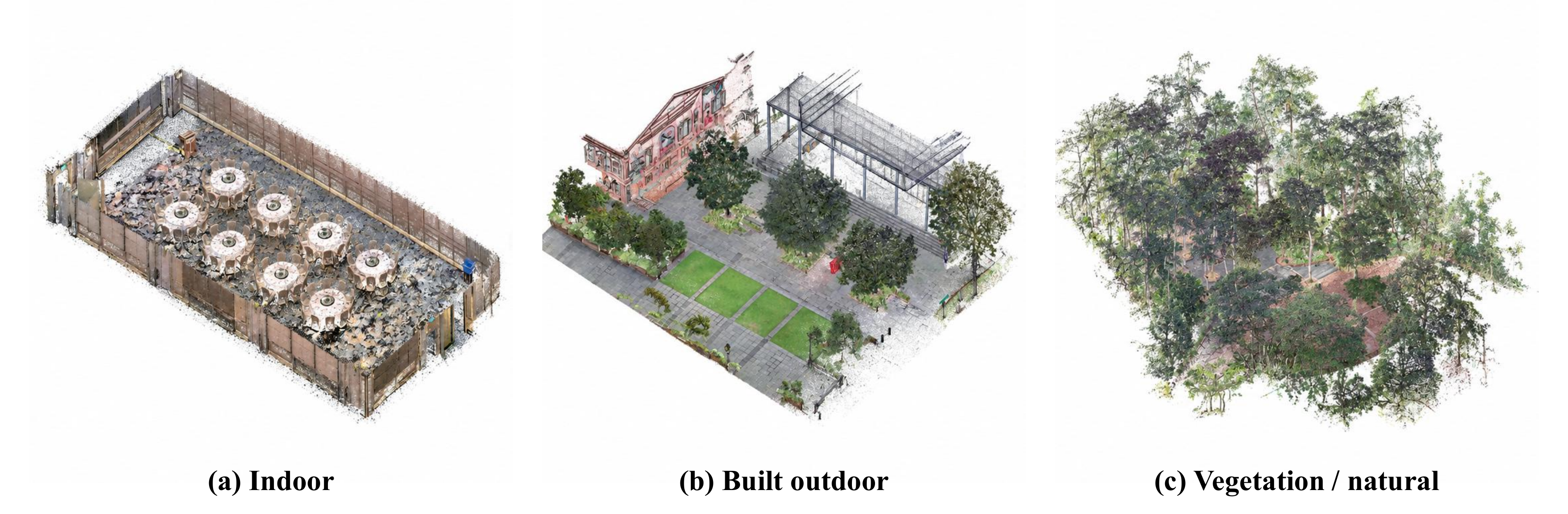}
\caption{\textbf{Reconstructed scenes across three environment categories.} Indoor, urban outdoor, and vegetation examples.}
\label{fig:app_scene_pointclouds}
\end{figure}

Figure~\ref{fig:app_scene_pointclouds} illustrates reconstructed scenes from the three categories.

\begin{table}[!ht]
\centering
\caption{{\AEheadingfont\textbf{Dataset splits.}}}
\label{tab:app_eval_scenes}
\small
\setlength{\tabcolsep}{3pt}
\begin{tabular}{@{}lrrrrrr@{}}
\toprule
& \multicolumn{4}{c}{Scenes} & Flight data & Area (m$^2$) \\
\cmidrule(lr){2-5}\cmidrule(lr){6-6}
Split & Indoor & Urban outdoor & Vegetation & Total & Trajectories & Min--max \\
\midrule
Training & 11 & 38 & 21 & 70 & 70{,}000 & 23--6{,}674 \\
Validation & 3 & 3 & 2 & 8 & 8{,}000 & 14--1{,}640 \\
Unseen test & 4 & 7 & 5 & 16 & 16{,}000 & 118--4{,}543 \\
\midrule
Total & 18 & 48 & 28 & 94 & 94{,}000 & 14--6{,}674 \\
\bottomrule
\end{tabular}
\end{table}

\paragraph{{\AEheadingfont\textbf{Evaluation episodes.}}}
The full closed-loop evaluation protocol specifies 16 seen scenes from the training partition and all 16 unseen test scenes, with 50 fixed episodes per scene (800 per split; 1,600 per policy and navigation setting). The seen subset covers the three environment categories and a range of spatial structures and areas (Table~\ref{tab:app_eval_scene_list}). Exact scene IDs, episode configurations, and sampling seeds will accompany the dataset release. Episode sampling follows the same scene-adaptive procedure used for data collection (Sec.~3.3). The protocol fixes scene IDs, initial poses, goals, geometry, and task seeds across policies and navigation settings; goal conditioning changes between settings. Seen evaluation tasks are held out from training demonstrations. Evaluation uses the scenes' existing static obstacles, without adding synthetic obstacles.

\begin{table}[H]
\centering
\caption{{\AEheadingfont\textbf{Evaluation scenes.}}}
\label{tab:app_eval_scene_list}
\small
\setlength{\tabcolsep}{3pt}
\renewcommand{\arraystretch}{1.08}
\begin{tabular}{@{}p{.16\linewidth}p{.285\linewidth}crcr@{}}
\toprule
& & \multicolumn{2}{c}{Seen} & \multicolumn{2}{c}{Unseen} \\
\cmidrule(lr){3-4}\cmidrule(lr){5-6}
Environment & Spatial structures & Scenes & \shortstack{Area range\\(m$^2$)} & Scenes & \shortstack{Area range\\(m$^2$)} \\
\midrule
\raggedright Indoor & \raggedright Rooms, corridors, open halls & 4 & 80--1{,}079 & 4 & 118--863 \\
\raggedright Urban outdoor & \raggedright Covered passages, streets, plazas & 7 & 200--6{,}674 & 7 & 502--4{,}543 \\
\raggedright Vegetation & \raggedright Wooded paths, hillside gardens, coastal parks & 5 & 141--1{,}191 & 5 & 128--1{,}678 \\
\bottomrule
\end{tabular}
\end{table}

\paragraph{{\AEheadingfont\textbf{Scaling evaluation.}}}
The scaling protocol retains the five seen scenes shared by the 5-, 15-, 35-, and 70-scene training subsets and uses five fixed unseen test scenes. Each scene contributes 100 fixed episodes: 500 seen and 500 unseen per training seed. The same tasks are reused across training scales and the three seeds. Metrics are averaged equally over scenes within each split for each seed. The figure reports individual seed results; the main-text endpoint values refer to the median across the three seeds. Scaling therefore uses a different scene subset and episode count from the main benchmark.

\paragraph{{\AEheadingfont\textbf{Termination.}}}
An episode succeeds when the drone enters the goal's 0.45\,m horizontal radius without a safety violation; no stop action is required. Failure occurs on collision, leaving the navigable polygon, or reaching the 5{,}000-step limit (100\,s). Collision is detected when obstacle distance in the scene-aligned 3D distance field is at most the 0.12\,m body radius (Sec.~3.2).

\paragraph{{\AEheadingfont\textbf{Offline metrics.}}}
We select 20 held-out trajectories of at least 152 steps per scene by a fixed seeded hash ordering (seed 20260910). All policies share up to six distinct expert anchors per trajectory: steps 0 and 50 (0 and 1\,s), and the rounded 10/35/65/90\% positions within $[101,T-51]$, where $T$ is the trajectory length in 50\,Hz steps. Duplicate indices after rounding are removed, and all policies use the same valid anchors. This retains at least 1\,s of future data at each anchor. Predicted and expert trajectories share the initial pose and are compared on a $\Delta t=0.02$\,s grid. For $K=h/\Delta t$ future positions $\hat{\mathbf p}_k,\mathbf p_k\in\mathbb R^2$,
\begin{equation}
\mathrm{ADE}_h=\frac{1}{K}\sum_{k=1}^{K}\|\hat{\mathbf p}_k-\mathbf p_k\|_2,
\qquad
\mathrm{FDE}_h=\|\hat{\mathbf p}_K-\mathbf p_K\|_2.
\end{equation}
We use $h=0.2$\,s. Control predictions are integrated into kinematic trajectory proxies; waypoint positions are linearly interpolated between native timestamps onto the evaluation grid. Additional queries within the window use recorded expert observations rather than observations from predicted states. CNN-BC is queried at every expert state in the window. For OmniVLA and NavMorph, the 0.2\,s window joins eight controls from the initial query with two from a new expert-state query at step 8; missing query coverage is unavailable. These are teacher-forced offline predictions. Table~\ref{tab:app_offline_metrics} defines additional diagnostics over the same 0.2\,s window, with subsets determined by expert behavior. Position errors are computed in meters and converted to centimeters for display in Table~\ref{tab:policy_benchmark}; angles are in radians, and lower errors are better.

\begin{table}[H]
\centering
\caption{{\AEheadingfont\textbf{Offline diagnostics.}}}
\label{tab:app_offline_metrics}
\small
\setlength{\tabcolsep}{4pt}
\begin{tabular}{p{.30\linewidth}p{.65\linewidth}}
\toprule
Metric & Definition \\
\midrule
Motion Direction Error & Angle between predicted and expert net displacements; undefined for zero displacement. \\
Band-limited DTW & Minimum cumulative position distance over monotone alignments, divided by $K$; band $|i-j|\leq\max(1,\lceil0.1K\rceil)$. \\
Lateral / Yaw Direction Error Rate & Opposite-sign fraction for expert lateral displacement $|d_y|\geq0.02$\,m / yaw change $|\Delta\psi|\geq0.02$\,rad. Lateral displacement uses the initial body frame. \\
Yaw Error & Absolute wrapped difference between predicted and expert yaw changes at the 0.2\,s endpoint, averaged over valid anchors within trajectories, then equally across trajectories within scenes and across scenes. \\
P95 FDE & 95th percentile of anchor-level FDE within each scene, then averaged equally across scenes. \\
Corner FDE & FDE for a $\geq20^\circ$ angle between the first and last 0.1\,s displacements of the next second; both lengths must exceed 0.02\,m. \\
Turning-segment FDE & FDE with 0.2\,s expert history and 1\,s future, where $|\omega|\geq0.1$\,rad/s in either adjacent 0.2\,s window. \\
\raggedright Turn-onset / Turn-end Yaw Error & With 0.2\,s expert history and 1\,s future, yaw error at $|\bar\omega^-|<0.1\leq|\bar\omega^+|$ / $|\bar\omega^+|<0.1\leq|\bar\omega^-|$, where $\bar\omega^-,\bar\omega^+$ are mean expert yaw rates (rad/s) in the preceding/following 0.2\,s windows. \\
Start / Mid-trajectory FDE & FDE at $t<2$\,s / $t\geq2$\,s; mid includes all post-start states. \\
Maneuver-balanced FDE & Equal-weight mean of FDE across three expert-demand bins. \\
\bottomrule
\end{tabular}
\end{table}

Demand bins use the larger of the bins for mean absolute lateral speed and yaw rate over the next second. Cutoffs are the 50th and 90th percentiles of 4{,}000 training windows from four scenes: approximately $(5.858712\!\times\!10^{-5},0.03238329)$\,m/s and $(1.464765\!\times\!10^{-4},0.08099765)$\,rad/s, respectively. Values at a cutoff enter the lower bin. Each scene-level mean requires 5 valid samples, P95 requires 20, and balanced FDE requires 5 per bin; otherwise the value is missing.

Offline means are computed over valid anchors within each trajectory, then averaged equally across trajectories within each scene; split summaries weight scenes equally. Cross-policy correlations use one sample per policy after averaging over common scenes, so $n$ counts valid policies. Within-policy correlations instead use one sample per scene. Spearman's $\rho$ is the Pearson correlation of ranks, averaging tied ranks; constant inputs yield an undefined correlation. For offline error $e$ and deployment metric $y$, direction-aligned correlation is
\begin{equation}
\rho_{\mathrm{align}}(e,y)=
\begin{cases}
-\rho(e,y), & y\in\{\mathrm{SR},\mathrm{A}^{\ast}\text{-}\mathrm{SPL},\mathrm{OSR}\},\\
\phantom{-}\rho(e,y), & y\in\{\mathrm{NE},\mathrm{CR},\mathrm{OB},\mathrm{Timeout},\mathrm{TTS}\}.
\end{cases}
\end{equation}
Positive values indicate that larger offline errors accompany worse deployment outcomes; negative values indicate the reverse. Paired differences subtract the reference diagnostic's aligned correlation using identical policies, scene coverage, and deployment outcomes. Missing entries are excluded jointly from both correlations.

\paragraph{{\AEheadingfont\textbf{Execution characteristics.}}}
ControlVar is the mean Euclidean change between consecutive normalized control vectors after command limiting. Turn/m is the accumulated absolute change in planar path direction per meter traveled, measured in rad/m.

\paragraph{{\AEheadingfont\textbf{Closed-loop metrics.}}}
For $N$ completed episodes, SR, CR, OB, and timeout are the percentages ending in each corresponding outcome. NE is mean terminal 3D goal distance; OSR is the percentage whose trajectory ever enters the goal's 1\,m 3D radius. With success indicator $S_i$, executed horizontal path length $P_i$, reference length $L_i$ of the A* path after line-of-sight shortcutting, and control-step count $T_i$,
\begin{equation}
\mathrm{A}^{\ast}\text{-}\mathrm{SPL}=\frac{1}{N}\sum_{i=1}^{N}S_i\frac{L_i}{\max(L_i,P_i)},
\qquad
\mathrm{TTS}=\frac{\Delta t\sum_i S_i T_i}{\sum_i S_i}.
\end{equation}
A$^\ast$-SPL uses $[0,1]$; TTS averages simulation seconds over successes and is missing if none succeed. The reference ends at the goal point, whereas execution stops within its success radius. Thus $P_i\leq L_i$ can saturate the score at 1; A$^\ast$-SPL measures efficiency relative to the grid-based planner, not an exact continuous-space optimum. Under the full protocol, closed-loop metrics other than TTS are computed within each scene and then averaged equally across scenes within each split. With 50 episodes in every scene, this equals pooling episodes for SR, CR, OB, timeout, NE, and A$^\ast$-SPL. TTS remains pooled over successful episodes.

The matched perturbation protocol and its local safety metrics are defined in Appendix~\ref{app:local_geometry}.

\subsection{Baseline Configurations}
\label{app:baselines}
\label{app:policy_api}
\suppressfloats[t]

\paragraph{{\AEheadingfont\textbf{Shared interface.}}}
AirPlayground shares data, controls, and evaluation while preserving native policy interfaces. In $\tau=\{(o_t,s_t,g_t,a_t)\}_{t=1}^{T}$, $o_t,s_t,g_t,a_t$ denote RGB, flight state, goal, and recorded control $(u_x,u_y,\omega)$. The conversion from stored translational controls to physical velocity follows Appendix~\ref{app:collection} and the collection velocity scale; yaw is already in rad/s. Policy-native target normalization is separate from this storage convention. The training adapter constructs native inputs and targets: $(\mathbf x_t^m,\mathbf y_t^m)=\mathcal T_m(\tau,t)$.

At query $q$, $\mathcal O_m$ builds $\mathbf x_q^m$ from causal RGB/state/action histories, the goal, and policy memory, yielding $\hat{\mathbf y}_q^m=\pi_m(\mathbf x_q^m)$. Within the controller $\mathcal C_m$ of Sec.~3.4, decoder $\mathcal D_m$ produces a control sequence or metric plan and executor $\mathcal E_m$ applies it:
\[
\mathbf p_q^m=\mathcal D_m(\hat{\mathbf y}_q^m),\qquad
\mathbf u_t=\mathcal E_m(\mathbf p_q^m,s_t;\mathbf h_t^m,H_m),\quad q\leq t<q+H_m.
\]
Context $\mathbf h_t^m$ stores plans and policy memory. Execution uses state feedback where required and queries every $H_m$ steps. Horizon experiments fix weights, decoding, and history sampling while varying $H_m$. Query-triggered state updates, including NavMorph's CEM memory update, follow the resulting schedule.

\paragraph{{\AEheadingfont\textbf{Training.}}}
The API supports PointGoal, ImageGoal, and Goal-free conditioning where applicable; each adapter declares its supported modes. The main benchmark uses PointGoal: observations, state, and relative goals, without maps, A* paths, or simulator goal RGB. Evaluation follows Appendix~\ref{app:metrics}; pretraining, optimization, and inference compute are not equalized. CNN-BC, ACT, and DP initialize visual encoders from ImageNet; other methods use released weights. Tables~\ref{tab:app_policy_adapters}--\ref{tab:app_training_budgets} summarize adapters and source training configurations for the evaluated checkpoints.

\paragraph{{\AEheadingfont\textbf{Execution configuration.}}}
Execution settings use the official implementations as references where applicable, with adaptations to the shared flight-control interface. We explore nearby execution intervals around these adapted configurations on validation scenes and freeze the selected settings for evaluation. The horizons in Table~\ref{tab:app_policy_adapters} describe our deployed adapters rather than universal defaults of the original methods. The execution-horizon study varies these intervals separately with model weights fixed.

\begin{table}[t]
\centering
\caption{{\AEheadingfont\textbf{Policy adapters.}}}
\label{tab:app_policy_adapters}
\footnotesize
\setlength{\tabcolsep}{6pt}
\renewcommand{\arraystretch}{1.1}
\begin{tabular}{@{}lclccl@{}}
\toprule
Method & \shortstack{RGB\\history} & Native output & \shortstack{Pred. span\\(s)} & \shortstack{Exec.\\$H$} & Deployment adapter \\
\midrule
CNN-BC & 1/1 & 1 control & 0.02 & 1 & Direct / hold \\
ACT & 1/1 & 50 controls & 1.00 & 50 & Prefix \\
DP & 2/1 & 32 controls & 0.64 & 16 & Prefix \\
ViNT & 6/5 & 5 waypoints & 0.50 & 10 & Tracker \\
OmniVLA & 1/1 & 8 poses & 0.16 & 8 & Pose differences / prefix \\
NWM & 4/5 & Future view & 0.80 & 40 & Rank / hold \\
NavMorph & 30/1 & 8 poses & 0.16 & 8 & Pose differences / prefix \\
\bottomrule
\end{tabular}
\end{table}

RGB history denotes frame count/stride, including the current frame. Stride and execution horizon $H$ count 0.02\,s steps. Prediction span includes the final control interval for action outputs. Controls are clipped to $\pm2$\,m/s per planar axis and $\pm1$\,rad/s in yaw.

\begin{table}[t]
\centering
\caption{{\AEheadingfont\textbf{Training configurations.}}}
\label{tab:app_training_budgets}
\footnotesize
\setlength{\tabcolsep}{7pt}
\begin{tabular}{@{}lcccl@{}}
\toprule
Method & Updates & Batch & Peak LR & Training scope \\
\midrule
CNN-BC & 5,280 & 256 & (1--3)$\!\times\!10^{-4}$ & Full policy \\
ACT & 34,650 & 32 & $10^{-5}$ & Full policy \\
DP & 54,450 & 32 & $10^{-4}$ & Full policy \\
ViNT & 8,418 & 256 & $10^{-6}$--$10^{-4}$ & Goal modules + decoder \\
OmniVLA & 500 & 32 & (2--5)$\!\times\!10^{-5}$ & LoRA + adapters \\
NWM & 604 & 64 & $10^{-5}$ & Full predictor \\
NavMorph & 10,000 & 4 & $10^{-5}$ & RGB backbone + adapters \\
\bottomrule
\end{tabular}
\par\smallskip
\begin{minipage}{\linewidth}
Updates refer to the selected checkpoint's final training stage. Batch includes all GPUs and accumulation (NWM: contexts, four targets each). LR ranges span parameter groups; exact assignments and stage configurations accompany the checkpoints.
\end{minipage}
\end{table}

All methods use AdamW; \texttt{configs/baselines/} specifies schedules, weight decay, group LRs, and frozen modules. Earlier selected stages comprise 2,806 ViNT goal-adapter updates, 504 OmniVLA head/projector updates, and 302 NWM conditioning updates.

\paragraph{{\AEheadingfont\textbf{Direct-control policies.}}}
CNN-BC holds its predicted control until refresh. ACT and Diffusion Policy (DP) execute $H$-step prefixes and discard unused tails. DP uses 100 denoising steps per decision; ACT disables temporal ensembling by default.

\paragraph{{\AEheadingfont\textbf{Waypoint policies.}}}
Our ViNT adaptation adds learned point-goal conditioning and predicts five query-relative $(x,y,\cos\psi,\sin\psi)$ waypoints at 0.1\,s spacing and 0.2\,m XY scale. Its cached third waypoint defines a world-frame position/heading target; each control step divides body-frame position and wrapped heading errors by 0.3\,s. OmniVLA adds a pose-goal projector, LoRA, and eight-pose head. Both leave visual-goal slots blank in PointGoal mode.

NavMorph adds a goal projector and eight-pose head to its RGB backbone, uses aligned state/action histories, and resets CEM to its initial memory each episode. OmniVLA and NavMorph decode cumulative query-relative poses (0.1\,m XY scale) into control prefixes: difference consecutive poses, starting from the query-frame origin; rotate displacements into the preceding body frame; divide displacement and wrapped yaw changes by 0.02\,s and clip.

\paragraph{{\AEheadingfont\textbf{World-model policy.}}}
Our NWM PointGoal adapter converts body-frame goal $g$ into a terminal relative pose $(g_{\mathrm{local}},\theta)$, where $g_{\mathrm{local}}=g\min(1,1.6/\|g\|)$ and $\theta=\operatorname{atan2}(g_y,g_x)$. XY follows NWM's 0.17\,m normalization; yaw remains in radians. The temporal condition follows NWM's native normalization and corresponds to a 0.8\,s prediction horizon. Using observed RGB history, 10 DDIM steps generate this view. This view is model-generated, never rendered or retrieved from the simulator at the target location.

The 120 candidates comprise the mean control and 119 diagonal-Gaussian samples, clipped to $[-1,1]^3$. Mean and standard deviation are fixed from training-control statistics. Sampling uses a deterministic seed derived from the episode identifier and query index. NWM predicts 0.8\,s endpoint views with 10 DDIM steps and shared noise across candidates. LPIPS ranks them against the generated goal view; the best five controls are averaged and held for 40 steps.

New policies implement these adapters while reusing the simulator and evaluator.

\subsection{Additional Navigation Results}
\label{app:additional_navigation}
\suppressfloats[t]

\paragraph{{\AEheadingfont\textbf{Goal specifications.}}}
AirPlayground's unified API supports baseline training and evaluation with PointGoal, ImageGoal, and Goal-free conditioning. PointGoal supplies a relative target. Here, ImageGoal uses successive A$^\ast$ subgoal views, denoted \emph{Image Subgoal}; the API also supports a single terminal goal image. Goal-free withholds goal input and relative-goal state entries, retaining the target only for evaluation. Table~\ref{tab:app_goal_modes} reports navigation under these guidance settings.

Image subgoals are rendered at the episode's flight height along the A$^\ast$ route, sampled every 0.5\,m starting 0.5\,m from the start. For route length $L$, the final view is placed at arc length $\max(L/2,L-1\,\mathrm{m})$. Each view faces the next subgoal, and the final view faces the task goal. Before each policy step, the subgoal index advances monotonically while the current subgoal is within 3\,m Euclidean distance of the drone, stopping at the first farther subgoal or the final view. The selected image is supplied through the goal-image interface. This provides route guidance without requiring arrival at every subgoal.

\begin{figure}[t]
\centering
\includegraphics[width=\linewidth]{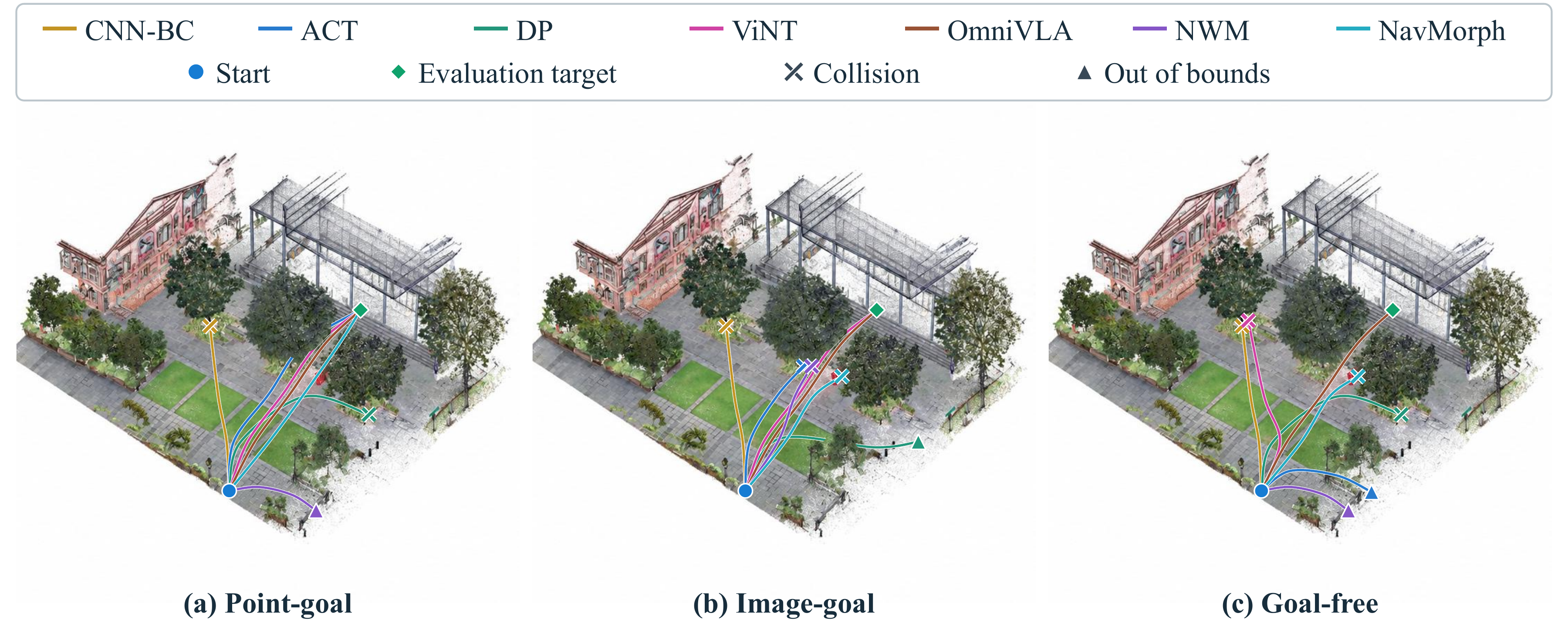}
\caption{\textbf{Navigation under different goal guidance.} Paths share a start and evaluation target; the target is withheld from Goal-free policies.}
\label{fig:app_goal_mode_trajectories}
\end{figure}

Figure~\ref{fig:app_goal_mode_trajectories} compares example paths under the three goal settings.

\paragraph{{\AEheadingfont\textbf{Shared evaluation tasks.}}}
The full three-setting protocol uses the same 16 seen and 16 unseen scenes and 50 start--goal episodes per scene for each supported policy--setting pair (800 per split; 1,600 total). PointGoal, ImageGoal, and Goal-free reuse initial poses, target locations, and scene geometry while changing goal conditioning. Except for TTS, metrics are first averaged within scenes, then equally across scenes within each split; a combined score gives seen and unseen equal weight. TTS pools successful episodes. Table~\ref{tab:app_goal_modes} reports navigation scores.

\begin{table*}[t]
\centering
\caption{{\AEheadingfont\textbf{Closed-loop navigation across goal modes.}} The full protocol reuses 50 tasks per scene across settings in 16 seen and 16 unseen scenes; see Appendix~\ref{app:additional_navigation} for aggregation and coverage.}
\label{tab:app_goal_modes}
\begingroup
\AEheadingfont
\scriptsize
\setlength{\tabcolsep}{2.6pt}
\renewcommand{\arraystretch}{1.12}

\resizebox{\textwidth}{!}{
\begin{tabular}{c *{7}{c} @{\hspace{5pt}} *{7}{c} @{\hspace{5pt}} *{7}{c}}
\toprule
\multirow{2}{*}{Method}
& \multicolumn{7}{c}{PointGoal}
& \multicolumn{7}{c}{ImageGoal}
& \multicolumn{7}{c}{Goal-free} \\
\cmidrule(lr){2-8}
\cmidrule(lr){9-15}
\cmidrule(lr){16-22}

& NE$\downarrow$ & OSR$\uparrow$ & SR$\uparrow$ & SPL$\uparrow$ & CR$\downarrow$ & OB$\downarrow$ & TTS$\downarrow$
& NE$\downarrow$ & OSR$\uparrow$ & SR$\uparrow$ & SPL$\uparrow$ & CR$\downarrow$ & OB$\downarrow$ & TTS$\downarrow$
& NE$\downarrow$ & OSR$\uparrow$ & SR$\uparrow$ & SPL$\uparrow$ & CR$\downarrow$ & OB$\downarrow$ & TTS$\downarrow$ \\
\midrule

CNN-BC
& 6.44 & 20.8 & 18.0 & 0.18 & 48.2 & 33.8 & 7.21
& 5.60 & 10.2 & 8.3 & 0.07 & 59.2 & 32.5 & 5.49
& 5.84 & 10.5 & 6.9 & 0.06 & \textbf{19.8} & 73.3 & 5.46 \\

ACT
& 4.38 & 39.1 & 35.1 & 0.35 & 52.1 & 12.7 & 4.74
& 4.08 & 26.7 & 24.8 & 0.22 & 54.3 & 20.9 & 3.21
& 4.34 & 16.4 & 12.1 & 0.10 & 63.5 & \textbf{24.4} & 2.42 \\

DP
& 7.17 & 26.5 & 15.9 & 0.14 & 57.1 & 27.0 & 5.88
& 4.13 & 24.4 & 20.2 & 0.16 & 52.7 & 27.1 & 3.58
& 4.43 & 16.4 & 13.1 & 0.11 & 60.1 & 26.8 & 3.75 \\

ViNT
& 4.94 & 39.0 & 35.9 & 0.36 & 50.4 & 13.7 & 8.05
& 2.20 & 52.1 & 51.1 & \textbf{0.48} & \textbf{39.7} & 9.2 & 6.01
& 4.36 & 17.1 & 13.4 & 0.11 & 60.8 & 25.8 & 3.45 \\

OmniVLA
& 3.81 & 46.5 & 44.0 & 0.43 & 49.1 & \textbf{6.9} & 6.83
& \textbf{2.00} & \textbf{53.8} & \textbf{52.3} & 0.47 & 42.8 & 4.9 & 5.44
& 4.08 & \textbf{32.6} & \textbf{19.6} & \textbf{0.15} & 49.4 & 31.0 & 3.74 \\

NWM
& \textbf{2.94} & \textbf{50.0} & \textbf{44.3} & \textbf{0.44} & \textbf{40.9} & 14.8 & \textbf{3.92}
& 2.60 & 44.2 & 36.4 & 0.33 & 46.8 & 16.8 & 3.92
& 4.73 & 19.1 & 11.4 & 0.09 & 57.5 & 31.1 & 2.75 \\

NavMorph
& 5.17 & 36.5 & 29.0 & 0.28 & 52.9 & 18.1 & 5.19
& 2.57 & 30.1 & 28.8 & 0.25 & 71.2 & \textbf{0.0} & \textbf{2.57}
& \textbf{4.01} & 19.6 & 15.1 & 0.13 & 57.2 & 27.7 & \textbf{2.29} \\

\bottomrule
\end{tabular}
}
\endgroup
\end{table*}

\paragraph{{\AEheadingfont\textbf{Task completion and safety.}}}
These results demonstrate the API's support for learning and executing policies under different forms of goal guidance. Image Subgoal yields higher SR than PointGoal for DP, ViNT, and OmniVLA, but lower SR for the other four methods. Successive visual subgoals provide intermediate route guidance, yet their benefits depend on the policy. Withholding goal input lowers SR for all seven methods. Safety still requires examining multiple failure modes: Goal-free CNN-BC has 19.8\% CR but 73.3\% OB, showing that low collision rate alone does not imply reliable navigation.

\subsection{Offline-to-Closed-Loop Analysis}
\label{app:offline_closed_loop}

\subsubsection{Offline Diagnostics}
\label{app:offline_diagnostics}

\paragraph{{\AEheadingfont\textbf{Main-text aggregation.}}}
The main-text cross-policy analysis uses seven baselines. For each policy and metric shown in the main-text correlation figure, we first average over common valid scenes within each split, then take the arithmetic mean of the seen and unseen split scores and compute Spearman correlations across the seven resulting policy scores; we do not average split-specific correlations. The full protocol specifies 16 seen and 16 unseen scenes with 50 closed-loop episodes per scene. The sample unit is a policy, giving seven observations for cross-policy correlations. Direction alignment follows Appendix~\ref{app:metrics}. For collision-rate comparisons, the signed difference is $\Delta\rho=\rho(e,\mathrm{CR})-\rho(\mathrm{FDE},\mathrm{CR})$, evaluated on identical valid policies and scene coverage. This is a descriptive cross-policy association, not an estimate of predictive accuracy. The evaluation protocol is specified in Appendix~\ref{app:eval_protocol}.

\subsubsection{Execution Horizon}
\label{app:execution_horizon}

\paragraph{{\AEheadingfont\textbf{Main-text evaluation protocol.}}}
The main-text execution-horizon study uses five seen and five unseen scenes with 50 fixed episodes per scene, totaling 500 episodes per policy and horizon. All horizons reuse the same tasks, checkpoint, prediction length, decoder, and history sampling for each policy. Scores are averaged within scenes and then equally within each split; combined scores weight seen and unseen equally. Each control step corresponds to 0.02\,s.

\begin{figure}[htbp]
\centering
\includegraphics[width=\linewidth]{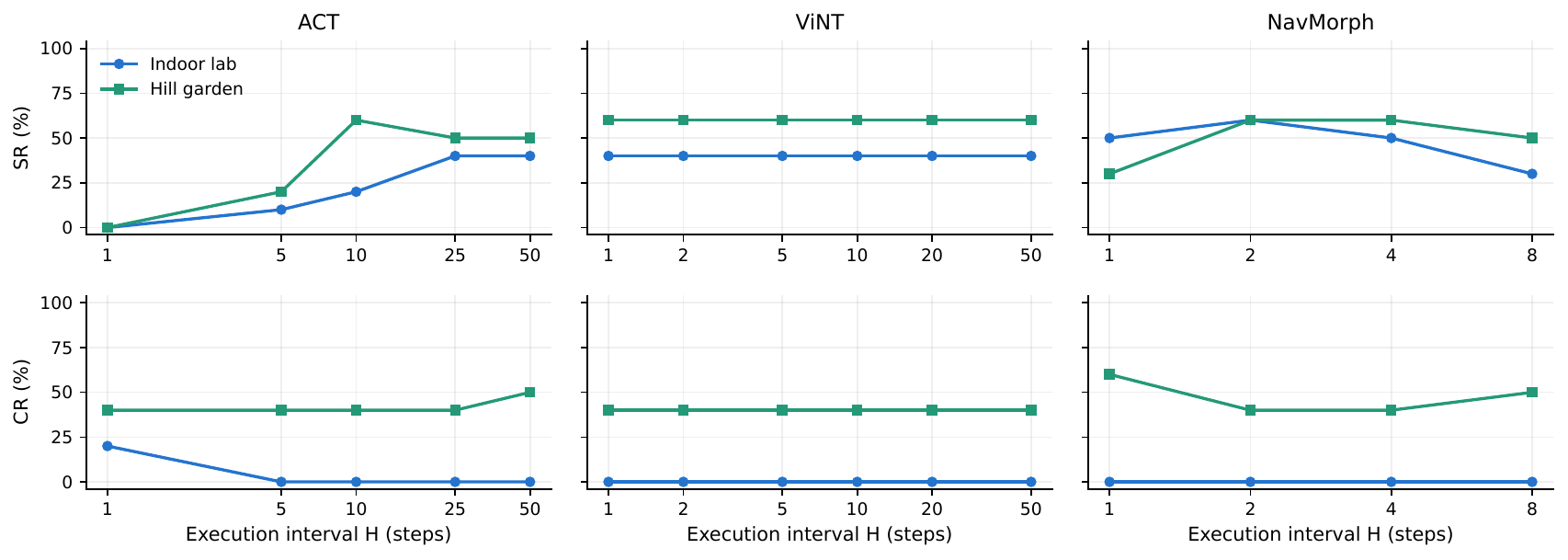}
\caption{\textbf{Execution-interval sweeps.} Supplementary two-scene evaluation, separate from the main-text sweep. Every point uses the same ten tasks in its scene; each method retains its checkpoint across intervals. The two rows show SR and CR. The horizontal axis lists all evaluated intervals on a logarithmic scale.}
\label{fig:app_horizon_sweep}
\end{figure}

\paragraph{{\AEheadingfont\textbf{Supplementary two-scene diagnostic sweeps.}}}
We vary $H$ on ten fixed tasks per scene in an indoor laboratory and a hill garden, retaining each checkpoint, prediction length, decoder, and history sampling (Fig.~\ref{fig:app_horizon_sweep}). Each step is 0.02\,s; execution follows Appendix~\ref{app:baselines}. ViNT retains its cached tracking target beyond the prediction span when $H$ is longer. Evaluation processes use seed 20260914, with one run per task and interval. ACT improves beyond $H=1$, NavMorph favors intermediate intervals, and ViNT retains the same SR and CR through $H=50$. These matched-task results show policy-dependent responses over the tested intervals.

\begin{figure}[htbp]
\centering
\includegraphics[width=\linewidth]{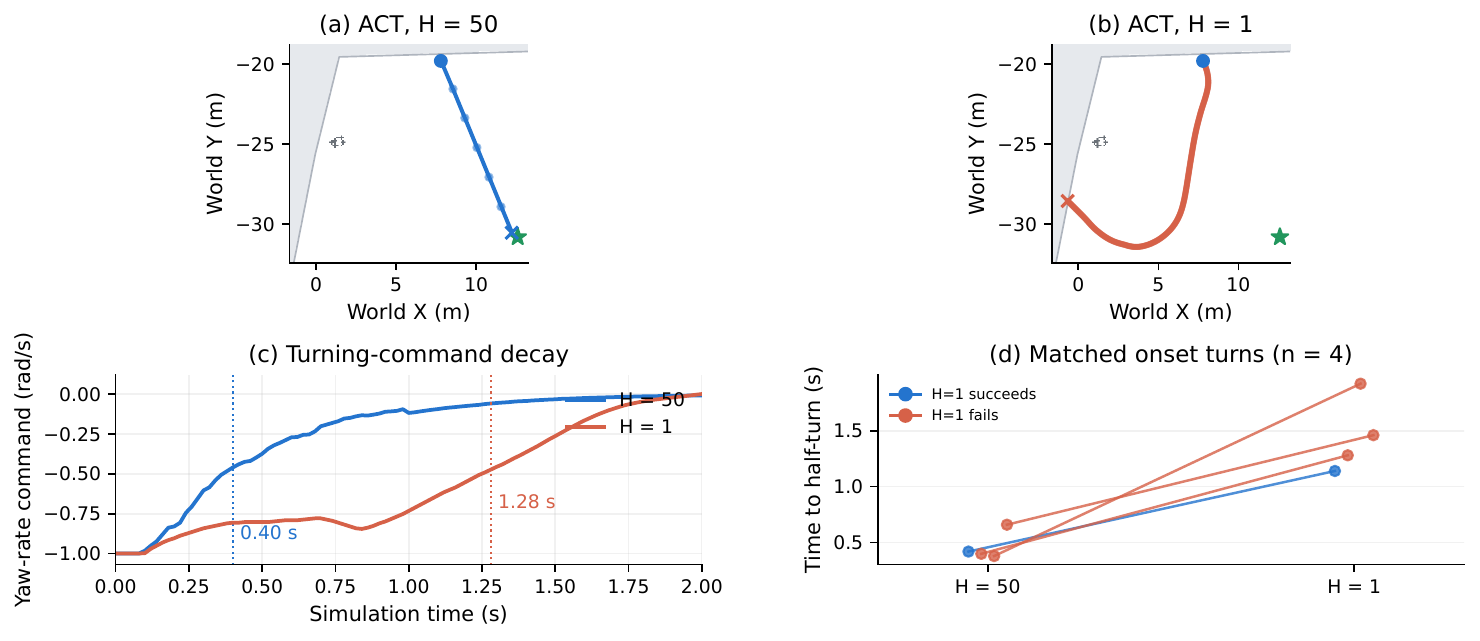}
\caption{\textbf{How execution changes a maneuver.} A separate matched-task ACT diagnostic. (a--b) Matched-start ACT trajectories on a flight-height occupancy slice: dark gray denotes obstacles and light gray lies outside the navigable polygon. Dots mark queries, stars goals, and crosses episode ends. (c) Executed yaw commands. (d) All four measurable turn-decay pairs, colored by $H=1$ outcome. States evolve independently after the shared start.}
\label{fig:app_execution_behavior}
\end{figure}

\paragraph{{\AEheadingfont\textbf{Turning behavior.}}}
A separate six-task ACT diagnostic compares $H=50$ and $H=1$ from matched starts. For each run, let $\bar\omega$ be the mean executed yaw-rate command over the first ten steps. We retain $|\bar\omega|\geq0.1$\,rad/s and define decay time as the start of the first subsequent five-step interval satisfying $\operatorname{sign}(\bar\omega)\omega_t\leq|\bar\omega|/2$ throughout. Four tasks have measurable decay in both runs; one lacks an initial turn and one terminates early. All four decay times increase under $H=1$, including 0.40 to 1.28\,s in the displayed task (Fig.~\ref{fig:app_execution_behavior}). This is a conditional maneuver diagnostic: one task still succeeds despite delayed decay, and early-terminated runs have no measured decay time.

\subsubsection{Clearance-Based Diagnostics}
\label{app:clearance_group_metrics}
Table~\ref{tab:geometry_clearance} summarizes 180 independent local rollouts across seven methods and ten scenes, with one rollout per sample. Starting states are selected near obstacles along expert trajectories, without conditioning on policy failures. At each start, the reference obstacle is selected in advance as the obstacle nearest to the expert's future path and remains fixed throughout the rollout, regardless of predictions or collision outcomes. Clearance is the distance from the drone center to the obstacle surface minus the body radius: positive values indicate remaining safety margin, and negative values indicate intrusion into the safety boundary. Expert and predicted minimum clearances refer to the same reference obstacle over a 0.2\,s prediction window (ten control steps at 50\,Hz); their difference is $\Delta c=c_{\min}^{\mathrm{pred}}-c_{\min}^{\mathrm{expert}}$. The decrease, stable, and increase groups use $\Delta c<-0.02$\,m, $|\Delta c|\leq0.02$\,m, and $\Delta c>0.02$\,m, yielding 43, 76, and 61 samples, respectively. FDE is the endpoint position error over that window. Distance columns report group means, with the closed-loop column averaging each rollout's local minimum clearance. Local passage requires reaching a predefined exit within 5\,s without collision or leaving the navigable region. The exit lies 0.5\,m beyond the reference obstacle's far boundary along the expert's direction of travel. Collision counts concern the reference obstacle; passage requires avoiding all collisions. Both counts use the full group size $N$ as their denominator, including other termination outcomes.

\clearpage
\subsubsection{Local Geometry}
\label{app:local_geometry}

\paragraph{{\AEheadingfont\textbf{Local intervention.}}}
ACT ($H=50$) and NavMorph ($H=8$) use nominal rollouts in an indoor laboratory and a courtyard. We select ten anchors per policy--scene--clearance group from collision-free states at step 30 or later, requiring speed 0.2--2.05\,m/s, goal distance at least 1.5\,m, and boundary clearance above 0.4\,m. Directions must pass a local SDF probe: moving $\pm0.2$\,m toward/away changes distance by at least 0.04\,m in the expected direction. Shared clearance cutoffs use quantiles of the four equally weighted candidate-frame pools. Branches share the recorded observation prefix and use live observations after intervention. We apply a world-frame planar velocity disturbance of 0.5\,m/s for up to 0.4\,s, stopping on episode termination. Toward uses the normalized XY vector $n=(n_x,n_y)$ from the anchor to its nearest occupied-voxel center; Away uses $-n$, and Tangential uses $\sigma(-n_y,n_x)$, with $\sigma\in\{-1,1\}$ fixed in the anchor manifest. Directions remain fixed during the pulse; the added increment is removed afterward. The disturbance is added after native velocity limiting without reclipping; it models an external disturbance, not a bounded policy output. Initial clearance subtracts the 0.12\,m body radius; shared bin boundaries are 0.520312 and 0.780000\,m. Directions are controlled within anchors; clearance bins compare different states.

\paragraph{{\AEheadingfont\textbf{Reference region.}}}
The reference patch comprises occupied voxels within 2\,m geodesic distance of the initially nearest occupied voxel, using edge-weighted 26-neighbor connectivity. All branches and strengths share an anchor-centered 1\,m-radius observation sphere. Observation ends at first exit, termination, or 5.4\,s; later events are excluded. Of 120 anchors, 93 have valid patches; 24 have isolated seeds and three have ambiguous nearest surfaces.

\paragraph{{\AEheadingfont\textbf{Local outcomes.}}}
Local CR is the reference-collision fraction over retained anchors, averaged equally across scenes; earlier non-reference terminations remain in the denominator without a reference-collision event. This is an observed local event fraction, not a censoring-adjusted risk. Minimum reference clearance $d_{\min}^{\mathrm{ref}}$ is the smallest native-grid distance to the reference patch minus body radius. Table~\ref{tab:app_local_geometry} uses branch-specific windows; paired clearance contrasts use the shortest window shared by all four branches. Intervals use 2{,}000 task-cluster bootstrap draws within scenes, preserving branch pairs and conditioning on these two scenes.

\begin{table}[htbp]
\centering
\caption{\textbf{Local disturbance outcomes.} Disturbance magnitude: 0.5\,m/s. $\mathrm{CR}^{\mathrm{ref}}$ is in percent and $d_{\min}^{\mathrm{ref}}$ in meters; $N$ counts accepted anchors. Results weight the two scenes equally.}
\label{tab:app_local_geometry}
\small
\setlength{\tabcolsep}{3.4pt}
\begin{tabular}{llr rr rr rr rr}
\toprule
& & & \multicolumn{2}{c}{Clean} & \multicolumn{2}{c}{Toward} & \multicolumn{2}{c}{Tangential} & \multicolumn{2}{c}{Away} \\
\cmidrule(lr){4-5}\cmidrule(lr){6-7}\cmidrule(lr){8-9}\cmidrule(lr){10-11}
Method & Margin & $N$ & $\mathrm{CR}^{\mathrm{ref}}\downarrow$ & $d_{\min}^{\mathrm{ref}}\uparrow$ & $\mathrm{CR}^{\mathrm{ref}}\downarrow$ & $d_{\min}^{\mathrm{ref}}\uparrow$ & $\mathrm{CR}^{\mathrm{ref}}\downarrow$ & $d_{\min}^{\mathrm{ref}}\uparrow$ & $\mathrm{CR}^{\mathrm{ref}}\downarrow$ & $d_{\min}^{\mathrm{ref}}\uparrow$ \\
\midrule
ACT & Low & 17 & 30.0 & 0.221 & 30.0 & 0.192 & 32.1 & 0.202 & 20.0 & 0.254 \\
ACT & Medium & 15 & 20.0 & 0.451 & 25.0 & 0.368 & 10.0 & 0.480 & 15.0 & 0.476 \\
ACT & High & 15 & 5.0 & 0.836 & 5.0 & 0.766 & 0.0 & 0.850 & 5.0 & 0.867 \\
\midrule
NavMorph & Low & 14 & 35.0 & 0.194 & 35.0 & 0.178 & 35.0 & 0.185 & 30.0 & 0.211 \\
NavMorph & Medium & 17 & 15.0 & 0.476 & 15.0 & 0.431 & 10.0 & 0.497 & 15.0 & 0.514 \\
NavMorph & High & 15 & 0.0 & 0.831 & 5.0 & 0.781 & 0.0 & 0.835 & 0.0 & 0.870 \\
\bottomrule
\end{tabular}
\end{table}

\begin{figure}[H]
\centering
\includegraphics[width=\linewidth]{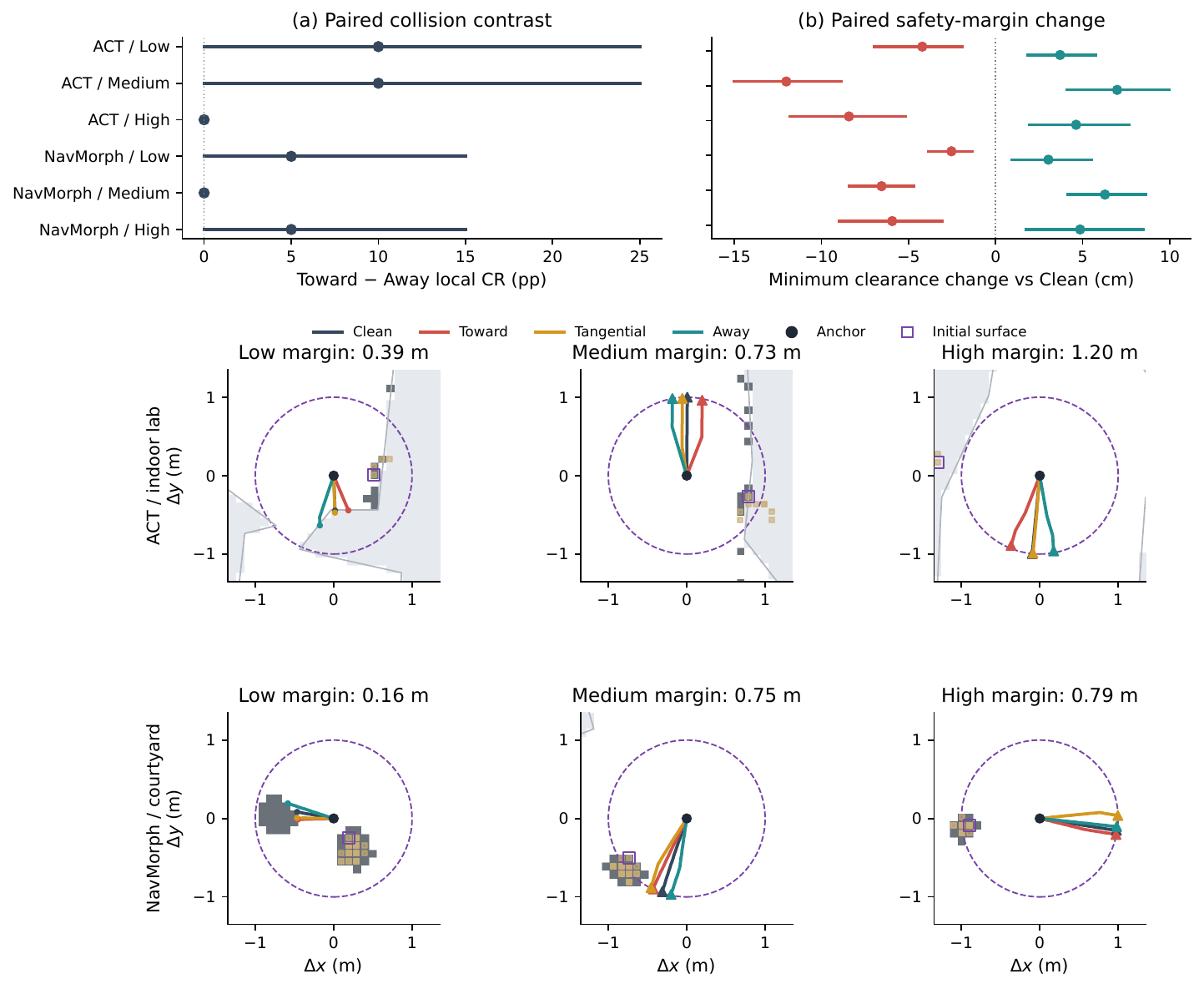}
\caption{\textbf{Paired effects within a fixed local region.} Controlled disturbances at 93 anchors, separate from the 180-sample clearance grouping. Top: local CR and common-window clearance contrasts with 95\% task-bootstrap intervals. Bottom: the first eligible anchor in each displayed policy--scene--clearance group. Dashed circles delimit the 1\,m observation region; tan cells show the reference patch. Crosses mark reference collisions and triangles region exits. Coordinates are anchor-relative. Zero-width CR intervals arise from all-zero paired differences and do not establish equivalence.}
\label{fig:app_local_geometry}
\end{figure}

\paragraph{{\AEheadingfont\textbf{Directional effects.}}}
Toward reduces common-window clearance relative to Clean and Away increases it in all six policy--bin groups (Fig.~\ref{fig:app_local_geometry}), while collision contrasts vary. On the same 46 NavMorph anchors, reducing the increment to 0.25\,m/s increases Toward clearance by 1.2/4.2/3.5\,cm in low/medium/high bins over windows common to both strengths, with unchanged local CR. Thus, binary collision counts can conceal a consistent loss of local safety margin.

\end{document}